\documentclass[letterpaper]{article} % DO NOT CHANGE THIS
\usepackage{aaai2026}  % DO NOT CHANGE THIS
\usepackage{times}  % DO NOT CHANGE THIS
\usepackage{helvet}  % DO NOT CHANGE THIS
\usepackage{courier}  % DO NOT CHANGE THIS
\usepackage[hyphens]{url}  % DO NOT CHANGE THIS
\usepackage{graphicx} % DO NOT CHANGE THIS
\usepackage{natbib}  % DO NOT CHANGE THIS AND DO NOT ADD ANY OPTIONS TO IT
\usepackage{caption} % DO NOT CHANGE THIS AND DO NOT ADD ANY OPTIONS TO IT
\usepackage{algorithm}
\usepackage{algorithmic}

\usepackage{newfloat}
\usepackage{amsmath}
\usepackage{booktabs}
\usepackage{multirow}
\usepackage{listings}
\DeclareCaptionStyle{ruled}{labelfont=normalfont,labelsep=colon,strut=off} % DO NOT CHANGE THIS
\floatstyle{ruled}
\newfloat{listing}{tb}{lst}{}
\floatname{listing}{Listing}
\title{KPLM-STA: Physically-Accurate Shadow Synthesis \\for Human Relighting via Keypoint-Based Light Modeling}
\author{
    Xinhui Yin\textsuperscript{\rm1,\rm3},
    ~Qifei Li\textsuperscript{\rm2,\rm3},
    ~Yilin Guo\textsuperscript{\rm4},
    ~Hongxia Xie\textsuperscript{\rm1,\rm2,\rm3}\equalcontrib,
    ~Xiaoli Zhang\textsuperscript{\rm1,\rm2,\rm3}\equalcontrib
    \\
}
\affiliations{
    \textsuperscript{\rm 1}College of Software Engineering, Jilin University, China\\
    \textsuperscript{\rm 2}College of Computer Science and Technology, Jilin University, China\\
    \textsuperscript{\rm 3}Key Laboratory of Symbolic Computation and Knowledge Engineering, Ministry of Education, Jilin University, China\\
    \textsuperscript{\rm 4}School of Computer Science, Peking University, China\\
}

\begin{document}

\maketitle

\begin{abstract}
Image composition aims to seamlessly integrate a foreground object into a background, where generating realistic and geometrically accurate shadows remains a persistent challenge. While recent diffusion-based methods have outperformed GAN-based approaches, existing techniques, such as the diffusion-based relighting framework IC-Light, still fall short in producing shadows with both high appearance realism and geometric precision, especially in composite images. To address these limitations, we propose a novel shadow generation framework based on a Keypoints Linear Model (KPLM) and a Shadow Triangle Algorithm (STA). KPLM models articulated human bodies using nine keypoints and one bounding block, enabling physically plausible shadow projection and dynamic shading across joints, thereby enhancing visual realism. STA further improves geometric accuracy by computing shadow angles, lengths, and spatial positions through explicit geometric formulations. 
Extensive experiments demonstrate that our method achieves state-of-the-art performance on shadow realism benchmarks, particularly under complex human poses, and generalizes effectively to multi-directional relighting scenarios such as those supported by IC-Light.
\end{abstract}

% Uncomment the following to link to your code, datasets, an extended version or similar.
% You must keep this block between (not within) the abstract and the main body of the paper.
% \begin{links}
%     \link{Code}{https://aaai.org/example/code}
%     \link{Datasets}{https://aaai.org/example/datasets}
%     % \link{Extended version}{https://aaai.org/example/extended-version}
% \end{links}

\section{Introduction}

Image composition \cite{composite} aims to combine foreground images and background images to generate composite images and has a wide range of applications. The problem addressed in this paper has broad practical application prospects, especially in the generation of composite image shadows in multiple scenarios and with multiple characters, such as virtual game scene transformation, obstacle shadow recognition in autonomous driving of automobiles, and estimation of building height based on shadows in remote sensing and geographic information systems, which are of great significance~\cite{intro_1,intro_2,intro_3,desobav2}. Therefore, this issue not only has academic research value but also has a promoting effect on practical application.
% \begin{textblock*}{5cm}(15.5cm,1cm)  % 宽度为5cm，位置(右边距5.5cm, 顶部1cm)
  
% \end{textblock*}
\begin{figure}[h]
    \centering
    \includegraphics[width=8cm,height=6cm]{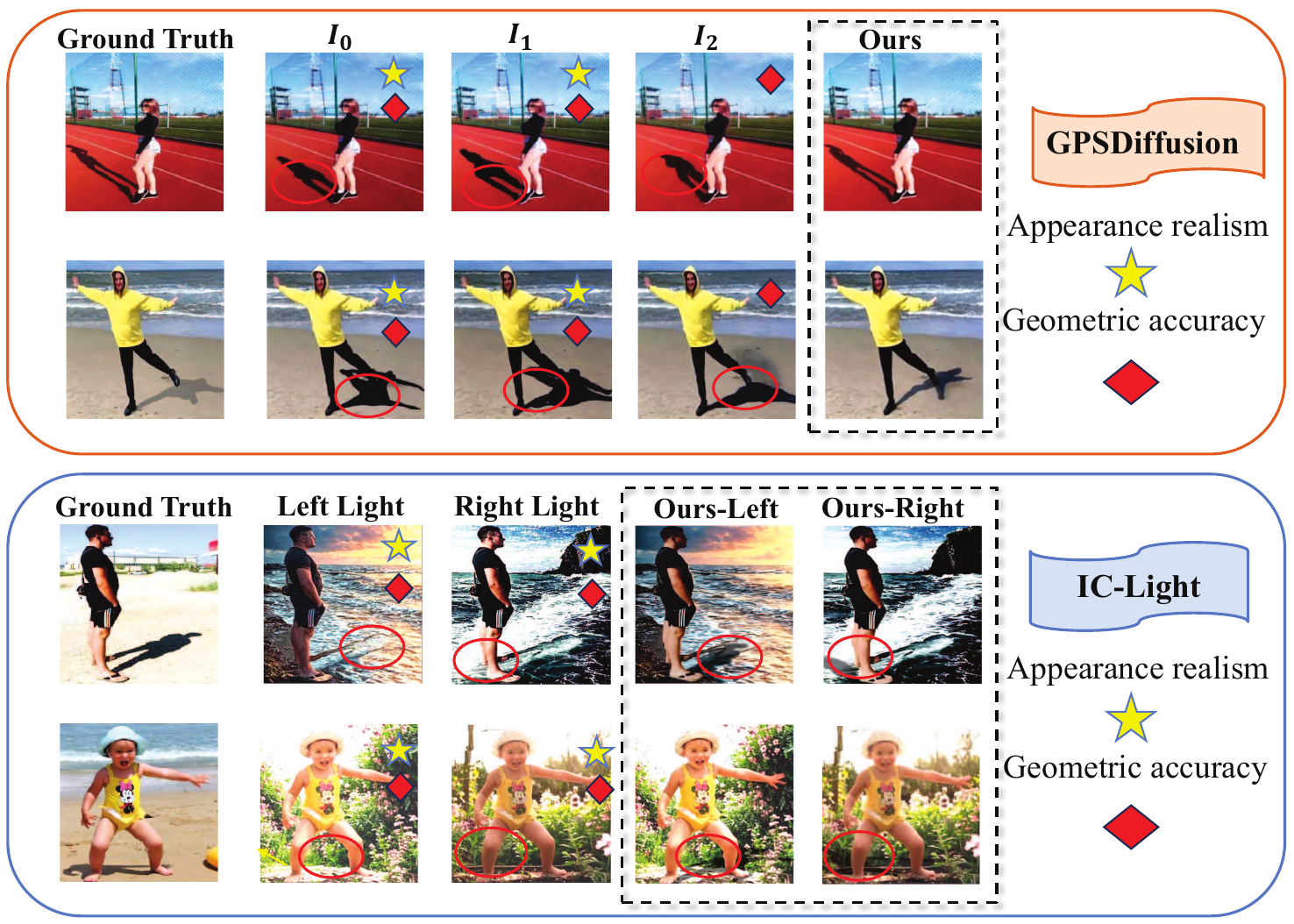}
    \caption{\textbf{Comparison with GPSDiffusion \cite{gpsdiff} and IC-Light~\cite{iclight}.}
The first row and columns $2-4$ are the random different results of GPSDiffusion, the second row and columns $2-3$ are the different light results of IC-Light. It can be seen that each image presents problems of appearance realism (shadow color is slightly darker) and geometric accuracy (limb shape is incorrect). However, our method (ours) has effectively solved these two problems.}
    \label{baseline}
\end{figure}

Diffusion-based methods~\cite{desobav2,gpsdiff} can leverage rich prior knowledge from pre-trained foundational models, consequently outperforming GAN approaches significantly and producing more realistic shadow effects.
Recent studies show that diffusion-based methods are effective in producing shadows with appearance realism, yet they struggle to capture geometric accuracy. As shown in Figure~\ref{baseline}, the problems of appearance realism and geometric accuracy are clearly visible. \textbf{Appearance realism} aims to make the generated shadows visually consistent with the physical properties of real shadows (such as color, transparency, softness and hardness, edge blurriness, etc.) and conform to the natural performance under lighting conditions, and \textbf{geometric accuracy} aims at the shape, position, and proportion of the shadow generated in three-dimensional space to be strictly consistent with the geometric relationships of objects and light sources in the scene. Imposing Consistent Light (IC-Light) \cite{iclight} has been established as a rapid diffusion-based composite image relighting technology. However, due to generating backgrounds via prompts before compositing with the foreground, IC-Light can preserve scene texture details after relighting, but still suffers from issues in appearance realism and geometric accuracy of ground shadows.

The task of generating shadows based on composite images has been a research focus in recent years. \textbf{Appearance realism} and \textbf{geometric accuracy} have always been the difficulties in the shadow generation task. To improve this problem, we propose KPLM-STA. The Keypoints Linear Model (KPLM) module precisely localizes key limb coordinates of the human body, ensuring the realism of shadow shape generation. Shadow Triangle Algorithm (STA) then estimates the corresponding shadow length and proportion for each limb based on varying viewpoints, light directions, poses, and ground angles. Through KPLM and STA, shadow angle, shape, length, and scale can be accurately controlled, resulting in realistic and geometrically consistent shadows.

% experiment 
% For shadow quality assessment, we primarily consider two dimensions:\textbf{ appearance realism} and \textbf{geometric accuracy}. 
 
 % \begin{figure}[htbp] \centering  %图片全局居中
 % \includegraphics[width=8cm,height = 5cm]{CameraReady/LaTeX/iclight.pdf} %\includegraphics[width=6cm,height = 4cm]{reduction_cell} 
 % \caption{Different Prompt and Light used IC-Light of the Image.} 
 % \label{iclight} 
 % \end{figure} 
 
In summary, our contributions are as follows:

\textbullet~We propose to equip diffusion-based shadow generation model with KPLM, which is a key point to enhance the apperance realism.

\textbullet~We introduce a shadow triangle algorithm to predict the shadow geometric for the foreground shadow.

\textbullet~We conducted experiments on the images processed by DESOBA \cite{desoba}, DESOBAv2 \cite{desobav2}, and IC-Light to verify the effectiveness of our method. Experimental results show that our method achieves the \textit{state-of-the-art} on three datasets, and it has a greater level of authenticity than previous methods.
\noindent

% \section{Copyright}
% All papers submitted for publication by AAAI Press must be accompanied by a valid signed copyright form. They must also contain the AAAI copyright notice at the bottom of the first page of the paper. There are no exceptions to these requirements. If you fail to provide us with a signed copyright form or disable the copyright notice, we will be unable to publish your paper. There are \textbf{no exceptions} to this policy. You will find a PDF version of the AAAI copyright form in the AAAI AuthorKit. Please see the specific instructions for your conference for submission details.

\section{Realted Work}
\subsection{Shadow Generation}

Shadow generation methods can be broadly categorized into\textbf{ GAN-based, geometry-guided, and diffusion-based paradigms}.
\textbf{GAN-based approaches}, such as Mask-ShadowGAN~\cite{mask-sg} and AR-ShadowGAN~\cite{ar-sg}, aim to synthesize visually realistic shadows via adversarial training. While effective in texture generation, they often suffer from structural inconsistencies—such as distorted contours or unnatural deformation under novel lighting—which limit geometric plausibility.

To address this, \textbf{geometry-guided methods }introduce structural priors to enforce alignment between objects and their shadows. DMASNet~\cite{dmasnet} and SGRNet~\cite{desoba} leverage bounding-box regression and physical illumination constraints to improve fidelity. However, these methods are primarily designed for rigid or simplified objects, making them unsuitable for articulated human figures.

Recently, \textbf{diffusion-based approaches} have gained attention for their flexibility and controllability. SGDiffusion~\cite{desobav2} and GPSDiffusion~\cite{gpsdiff} incorporate sketch priors or geometric embeddings to enhance spatial consistency. Yet, they largely focus on static objects or primitive shapes, without addressing the complexity of articulated human shadows. Moreover, most existing methods tend to optimize for either appearance realism or geometric accuracy—rarely both.

Meanwhile, \textbf{text-conditioned relighting frameworks }offer fine-grained lighting control. IC-Light~\cite{iclight} leverages a latent diffusion model to manipulate light direction and tone via prompts, achieving high-quality relighting with contextual consistency. However, it lacks explicit modeling of object-ground interaction and thus cannot generate cast shadows, undermining both photorealism and physical correctness. 
% Follow-up works such as SpotLight~\cite{spotlight} and Lite2Relight~\cite{ite2relight} improve local lighting with sketch or 3D priors, but similarly overlook cast shadow synthesis.

To address this gap, we extend IC-Light by explicitly modeling cast shadows for articulated human figures. Our method enables end-to-end generation of shadows that are both light-consistent and structurally grounded, advancing physically-aware, human-centric shadow synthesis.

\subsection{Human Model}

Human shadow generation requires structured body modeling. OpenPose~\cite{openpose,openposecpm,openposehand,openposerealtime} and MediaPipe~\cite{mediapipe} offer efficient 2D pose estimation from monocular images, providing keypoint landmarks that can represent body geometry. SMPL~\cite{smpl}, a parametric 3D mesh model, offers high-fidelity shape and pose reconstruction, but is often computationally intensive and unsuitable for lightweight relighting tasks. In our framework, we adopt a simplified 2D body abstraction using 9 keypoints and one bounding block, which captures sufficient spatial structure while enabling efficient shadow geometry inference. 

\begin{figure*}[h]
\centering  %图片全局居中
\includegraphics[width=15cm,height =8cm]{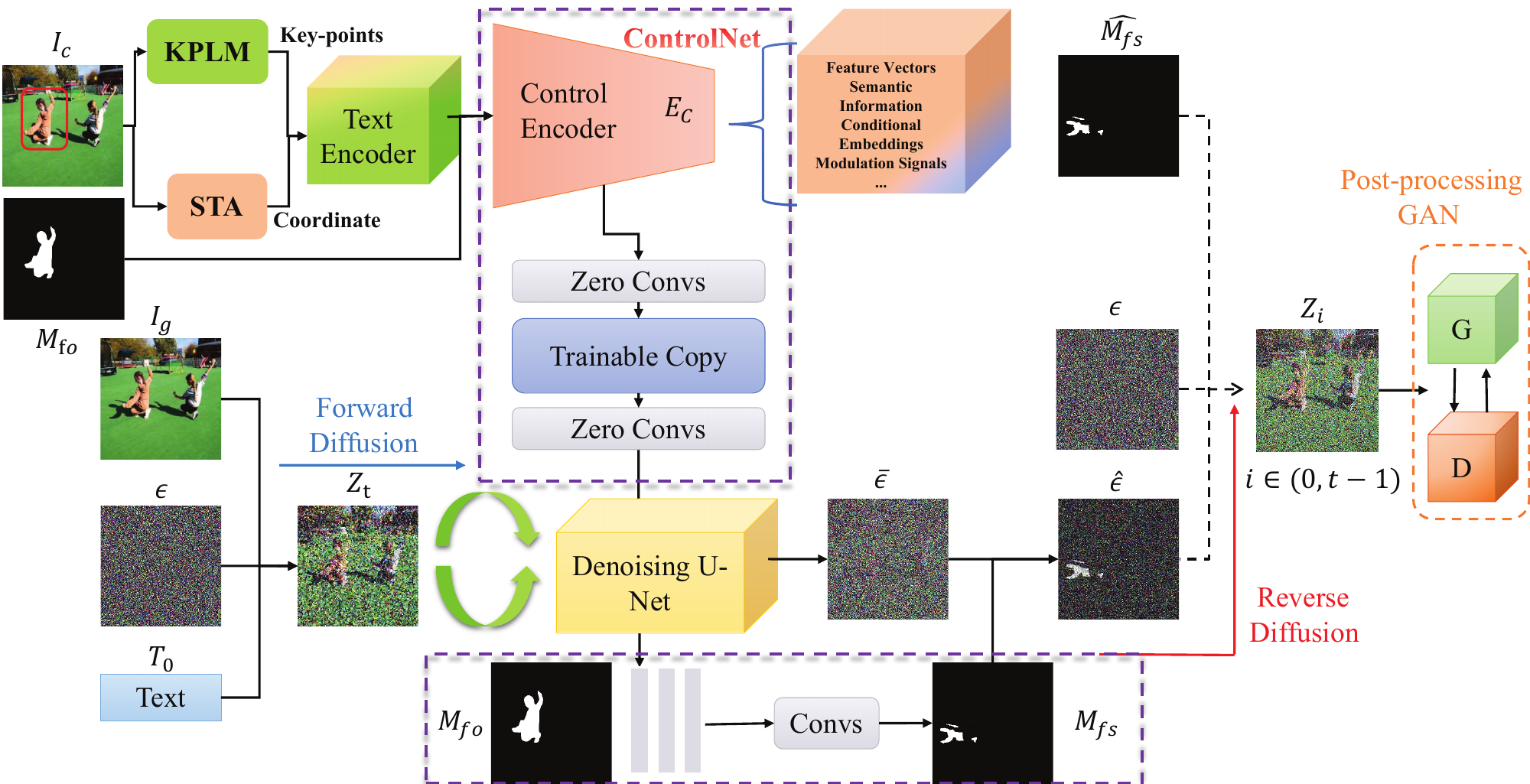}
\caption{\textbf{The framework of our KPLM-STA.} In the first stage, we use KPLM to get key point coordinates, use STA to obtain the shadow angle. In the second stage, we use Control Encoder and Diffusion to generate image. Finally, we use a post-processing GAN for realistic processing.}
\label{framework}
\end{figure*}

\section{Method}
Our method consists of three stages: \textbf{geometry-aware preprocessing}, \textbf{latent diffusion-based~\cite{ldm} shadow generation}, and\textbf{ GAN-based~\cite{gan} post-processing refinement}. Given a composite image $I_c$ without foreground shadow, we firstly extract geometric priors using Keypoints Linear Model (KPLM) and Shadow Triangle Algorithm (STA), which provide keypoint and directional shadow information. These priors are encoded and injected into a latent diffusion model (LDM)~\cite{ldm} via ControlNet~\cite{controlnet} following SGDiffusion~\cite{desobav2}. Finally, we employ a trainable GAN-based post-processing network to refine the generated image and alleviate color shift and background variation. An overview of our pipeline is shown in Figure~\ref{framework}.

\subsection{KPLM-STA}
\label{KPLM-STA}
In the first stage, KPLM and STA perform synchronized preprocessing to ensure geometric accuracy. As shown in Figure~\ref{framework}, both modules require two sets of computations. Given the input image $I_c$, KPLM determines the keypoint data of the foreground object through individual mapping of 9 key points and 1 key block, while STA models shadow triangles to estimate the overall position and length of the shadow, including missing limb parts. These geometric features are organized and encoded to control key factors such as shadow color and shape.

In the second stage, given the ground-truth shadow image $I_g$, noise $\epsilon$, and optional text input $T_0$, the latent representation is obtained via $Z_0 = E_r(I_g)$, and forward diffusion is computed by $Z_t = \epsilon + Z_0$. Following SGDiffusion, we apply ControlNet to inject spatial priors into the denoising U-Net. Specifically, the processed $I_c$ and $M_{fo}$ are concatenated and fed into the control encoder to produce conditional features, which are combined with $Z_t$ during denoising. Additionally, ControlNet uses $M_{fo}$ to predict the foreground shadow mask $M_{fs}$, which regulates the reverse diffusion process and guides the generation of intermediate latents $Z_i$. After decoding $Z_i$, the predicted shadow image $\hat{I}_g$ is obtained.

In the third stage, $\hat{I}_g$ is further refined by a multi-task GAN-based post-processing network that incorporates the foreground mask $M_{fo}$ and composite image $I_c$ to improve realism and remove color shift or background disparity.
\footnote{The technical details of post-processing GAN are left to Appendix.}

\subsection{Keypoints Linear Model (KPLM)}

\begin{figure}[htbp]
\centering  %图片全局居中
\includegraphics[width=6cm,height = 4cm]{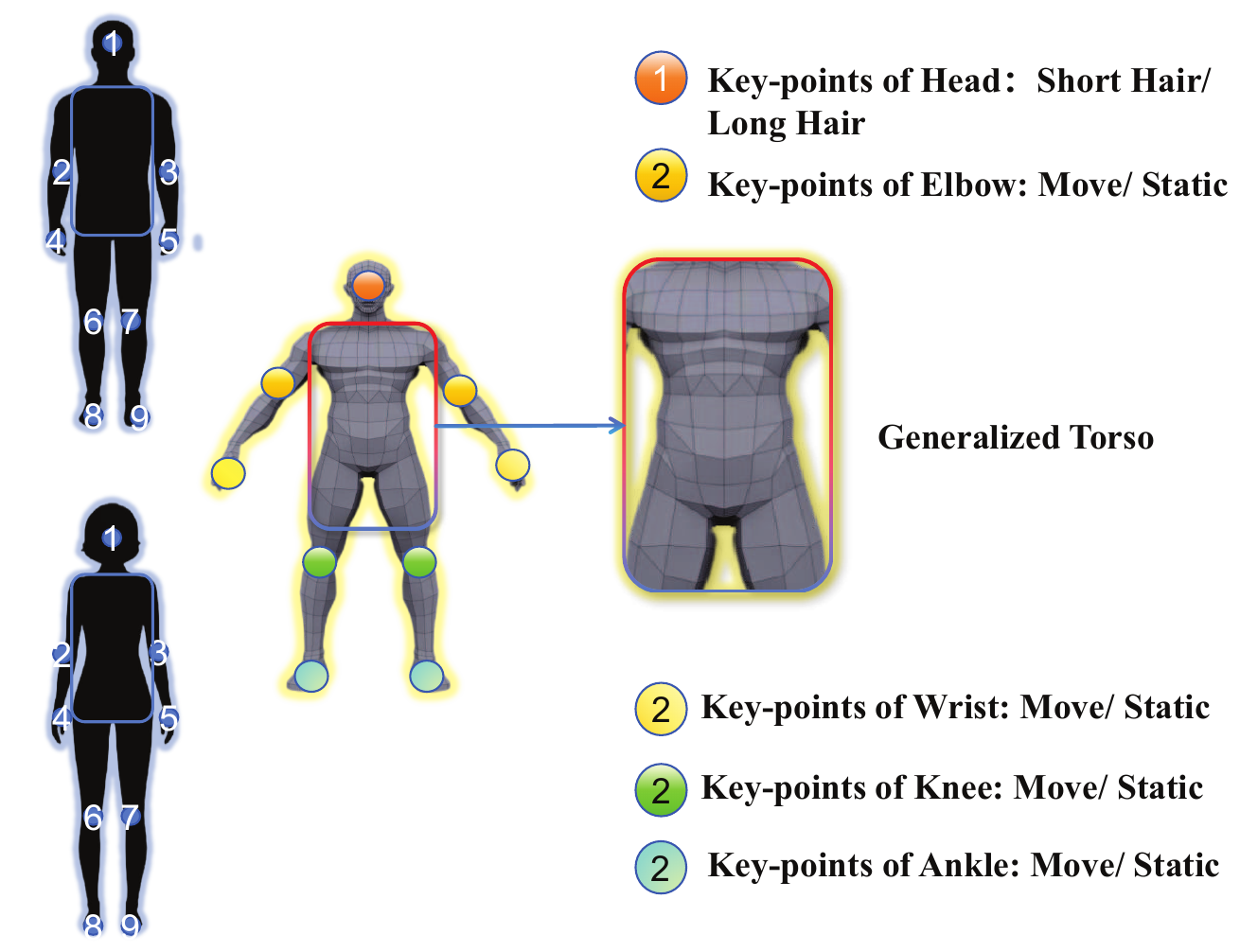}
\caption{KPLM modeling schematic diagram.}
\label{kplm}
\end{figure}

For the task of generating human body shadows in two-dimensional images, there are often problems such as missing limbs, incorrect positions and postures. Previous work such as Pose2Light~\cite{pose2light} estimated the lighting direction of human pose in coarse ambient scenes. We leverage 2D keypoints to model the light direction via a linear mapping. KPLM regards the trunk as a universal skeleton frame, which changes with human posture and angle. Unlike pose capture or other human-based tasks, the human body corresponding to the shadow does not require more detailed keypoint capture. KPLM can determine the general posture and position of the people in the picture, and it can achieve better results at a lower cost.
\begin{figure}[htbp]
\centering  %图片全局居中
\includegraphics[width=4cm,height = 3cm]{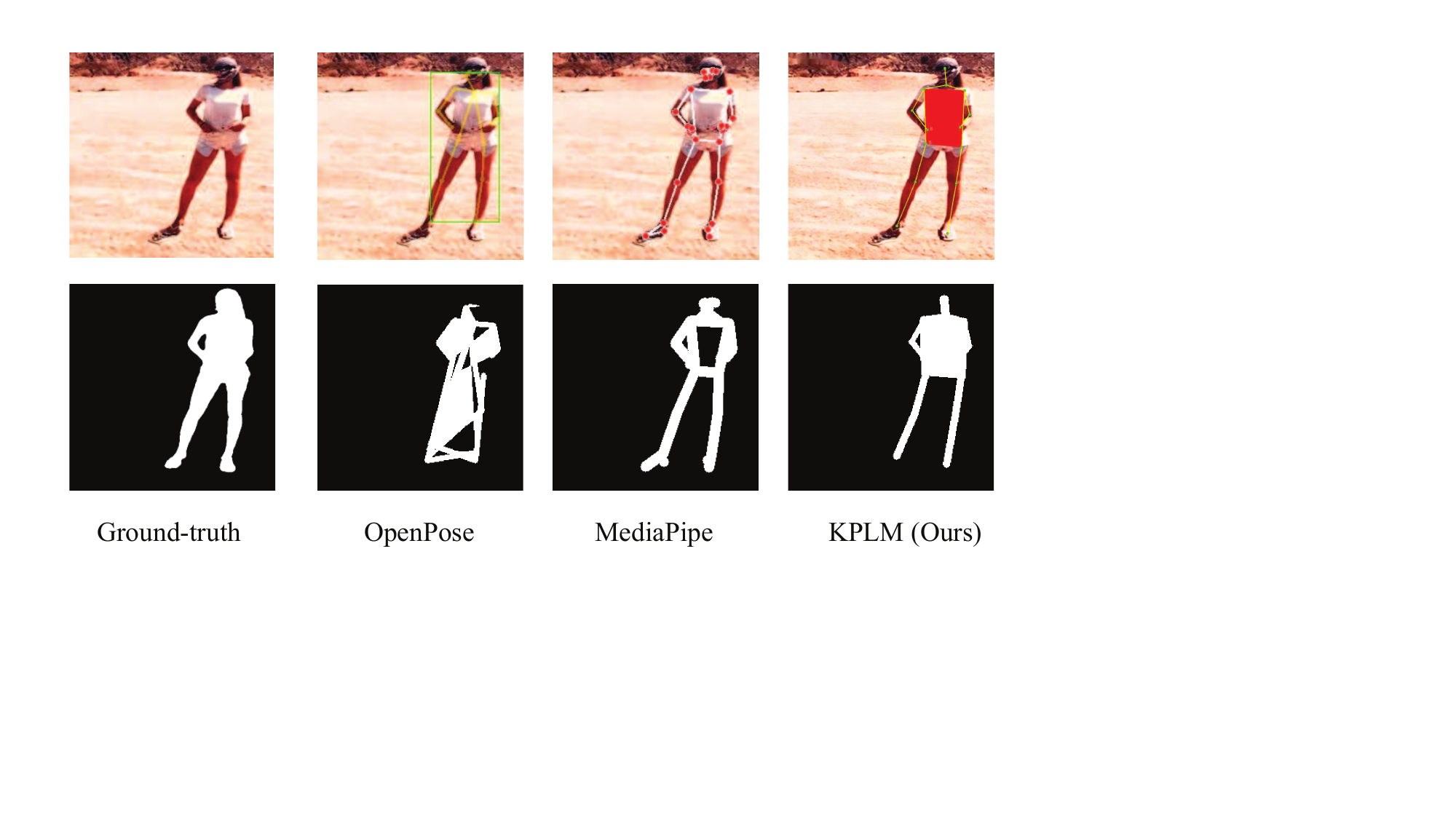}
\includegraphics[width=4cm,height = 3cm]{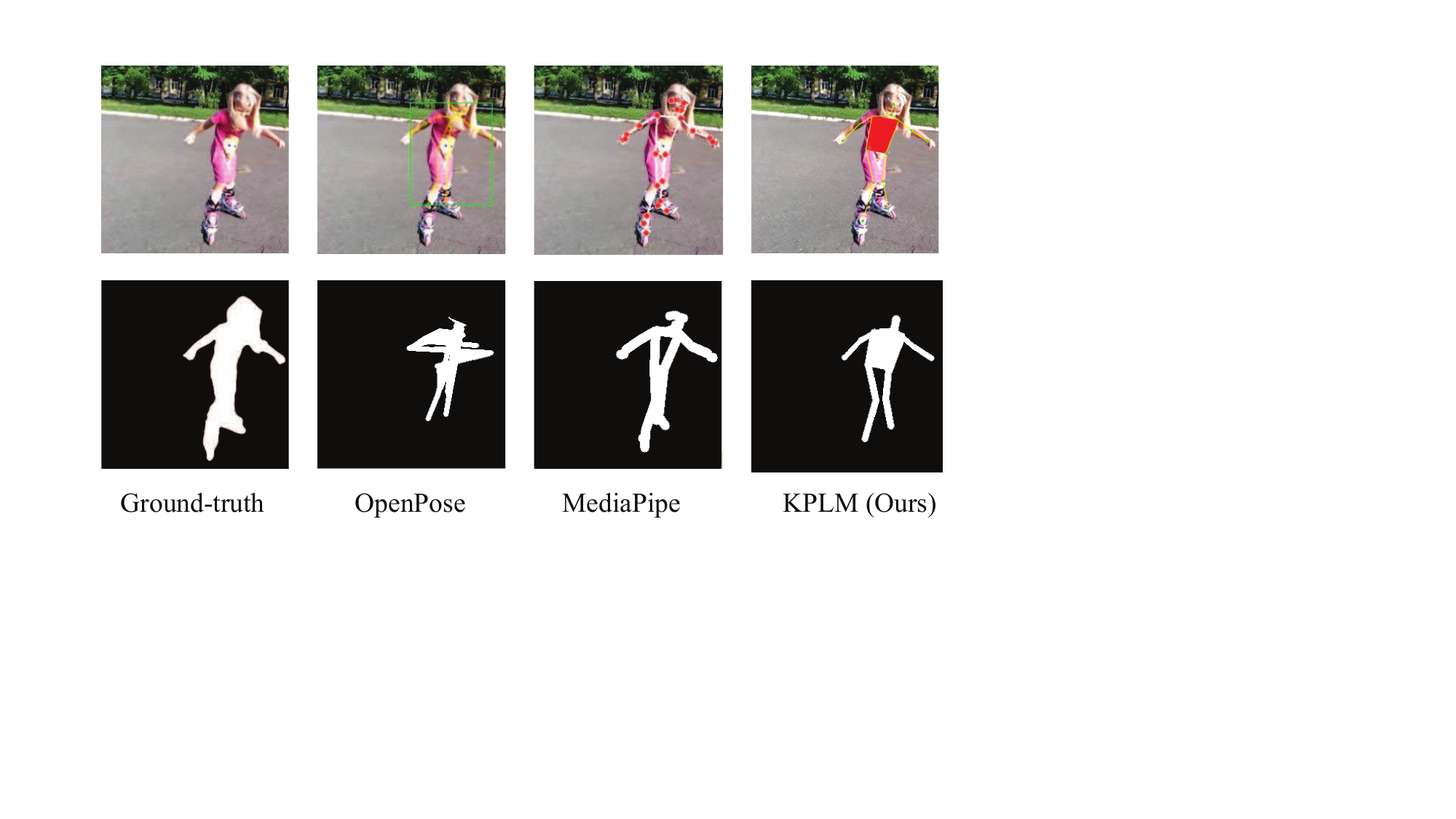}
\caption{KPLM compared with the mask of object in the most universal method OpenPose, MediaPipe.}
\label{mask}
\end{figure}

% \begin{table}[!ht]
% \centering
% \begin{tabular}{c|c|c|c}
% \hline
%     Model & IoU $\uparrow$ & Time (ms) $\downarrow$ & Numbers $\downarrow$ \\
% \hline
%     OpenPose & 0.411 & 5547 & 17 \\
% \hline
%     MediaPipe & \textbf{0.551} & 8207 & 33 \\
% \hline
%     \textbf{KPLM} & 0.535 & \textbf{5403} & \textbf{14}  \\ 
% \hline
% \end{tabular}
% \caption{Evaluation with OpenPose, MediaPipe on mask.}
% \label{kplm_mask}
%\end{table}

For human body postures, we adopt $9$ key points and $1$ key block for modeling. Moreover, we can abstract the changes in human shadow postures into the changes of $9$ key points as shown on the right side of Figure~\ref{kplm}, that is, the generation of human shadow is driven by $9$ key points. As shown in the figure, our $9$ key points are Head-$1$, Elbow-$2$, Wrist-$2$, Knee-$2$, Ankle-$2$ respectively, and the trunk can be regarded as a key block, which can be obtained through STA rotation scaling. As shown in Figure~\ref{mask}, our method was compared with the most universal methods, OpenPose and MediaPipe. The object mask of KPLM is closer to the real object mask than OpenPose, and KPLM uses fewer points than MediaPipe. 

In addition, the four vertices of the block change as the human body rotates, which ensures that the universal skeleton frame transforms as the human body rotates. The skeleton frame and key points of each character can be determined through KPLM to determine the human body posture and position. Although in our current implementation KPLM is treated as a pre-processing step, it can extracts structural pose representations from the input image. In this form, KPLM can be integrated into the full diffusion framework, enabling end-to-end training. We leave this extension for future work.\footnote{The module ablation experiment on KPLM (explaining why $9$ points $+$ $1$ block is optimal) will be presented in the second section of the Appendix.}

\subsection{Shadow Triangle Algorithm (STA)}

Prior work has explored physically-based shadow rendering by simulating light projection over estimated height maps, such as Pixel Height Maps~\cite{eccv}, which enable hard shadow synthesis under geometric constraints. However, such methods assume the human body stands vertically on a flat ground plane, limiting applicability to upright poses and restricting scene diversity. In contrast, STA supports arbitrary body positions and complex poses by enabling triangle-based shadow modeling for each limb individually.

Recent state-of-the-art methods like GPSDiffusion~\cite{gpsdiff} treat the human and its shadow as two separate bounding boxes, deriving the shadow via rotation and translation. While effective at a coarse level, this lacks the granularity needed for fine structural alignment. STA overcomes this by leveraging keypoints extracted from KPLM to perform triangle-based limb-level shadow projection, allowing precise control over the orientation and length of each limb’s shadow.

\begin{figure}[!ht]
\centering
\includegraphics[width=8cm,height = 3cm]{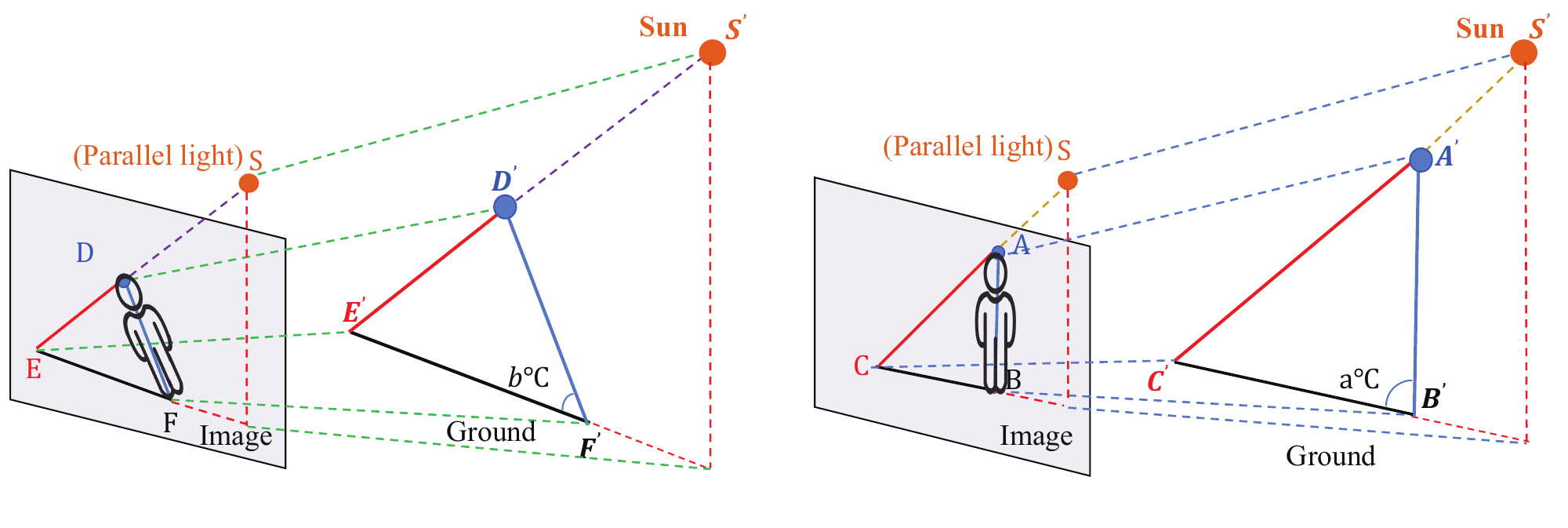}
\caption{\textbf{Shadow Triangle Construction.} Given a human limb segment AB and the corresponding shadow endpoint $C$, the triangle $\triangle ABC$ is formed under parallel light from the sun. $A'$ and $B'$ denote projected shadow points on the ground. The angle $\theta$ is formed between the light direction and the ground plane.}
\label{sta}
\end{figure}

\begin{figure}[!ht]
\centering
\includegraphics[width=8cm,height = 5cm]{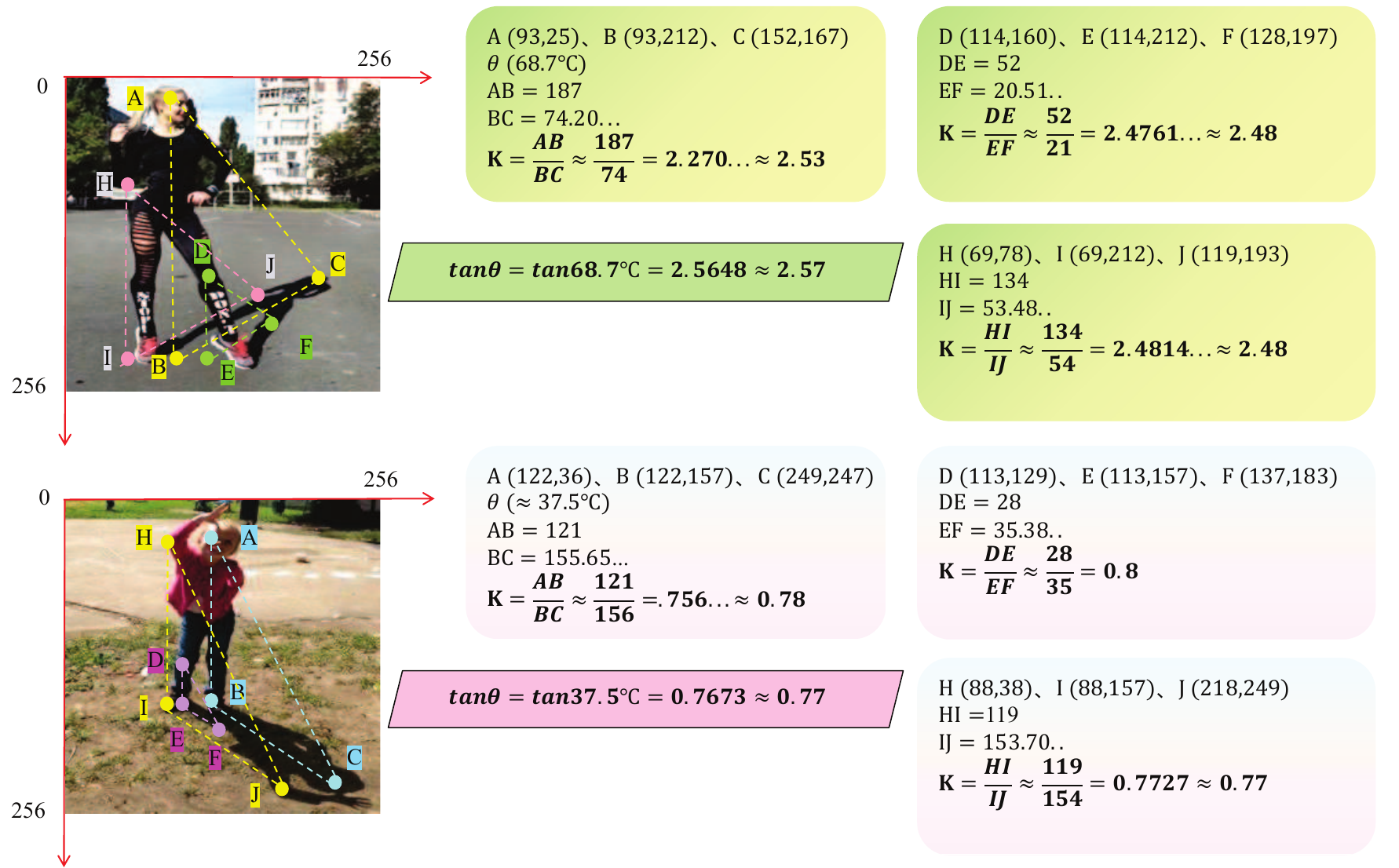}
\caption{\textbf{Proof of STA in Physical Formulas.} $K$ is the proportionality coefficient between body segments and shadows; $\theta$ is the angle between the light and the ground.}
\label{sta_1}
\end{figure}

STA models the person and their shadow as geometric segments. According to the principle of light projection, the light source direction is uniquely determined by the alignment of the person and the shadow. For points $A$ and $D$, let $C_1$ and $C_2$ be the corresponding circles of their heads, with centers $r_1$ and $r_2$, respectively by 
$\mathrm{Ar}_1\perp\mathrm{CS}, \quad \mathrm{Dr}_2\perp\mathrm{ES}$. The triangle formed by light direction, body, and shadow is called a shadow triangle. As shown in Figure~\ref{sta}, triangles $\triangle ABC$ and $\triangle DEF$ correspond to different light angles $a^\circ$ and $b^\circ$. The limbs and their shadows obey a linear scaling relationship that

\begin{equation}
\mathrm{BC=K \cdot AB}, \quad \mathrm{FE=K \cdot DF}.
\end{equation}

Since sunlight can be treated as parallel light, the shadow length of each limb is computed by scaling the limb vector with a coefficient $K$. To model this process differentiably, we formulate STA as an affine geometric projection. For a limb defined by two keypoints $P_1, P_2 \in \mathrm{R}^2$ on the image plane (e.g., elbow to wrist), we aim to compute its projected shadow endpoint $P_3$, forming a proportional relationship:

\begin{equation}
    \mathrm{K = \frac{\|P_1 - P_2\|_2}{\|P_2 - P_3\|_2}
},
    \label{EQ5}
\end{equation}
where $\|\cdot\|_2$ denotes Euclidean distance. This coefficient $K$ encodes the ratio between the limb and its shadow, varying with light direction, body orientation, and camera viewpoint. Under orthographic projection and directional lighting, we approximate $K \approx \tan \theta$,
where $\theta$ is the angle between the light and the ground. Given $K$, we compute the projected shadow position using:

\begin{equation}
\text{ShadowPos} = P_2 + \frac{1}{K} \cdot \|P_1 - P_2\|_2 \cdot 
\begin{bmatrix}
\cos(\theta) \\
\sin(\theta)
\end{bmatrix}.
\label{EQ6}
\end{equation}

Eq.~\ref{EQ6} simulates the projection of a limb along the light direction. After computing projected shadow endpoints for all limbs, we construct shadow limbs by connecting each original limb to its corresponding shadow. These projected limbs form the basis of the shadow mask used in the reverse diffusion process.

To validate STA, we examine both image-space consistency and physical correctness. First, we compare the scaling factor $K$ across the full human body, left elbow, and right knee in 2D image space. Results show all values of $K$ are approximately equal, with a maximum deviation of 0.05 due to keypoint approximation and sampling noise. Similarly, for another sample, the deviation among $K$ values is below $0.02$.

Second, in the physical world, shadow formation obeys 
$\tan \theta = \frac{h}{L}$, where $h$ is the object height and $L$ is shadow length. We empirically verify that $K \approx \tan \theta$ by extracting keypoint coordinates, computing limb and shadow lengths, and measuring $\theta$. As shown in Figure~\ref{sta_1}, the real-world angle $\theta$ aligns with that in the image, confirming that STA faithfully maps 3D shadow behavior to 2D projection, with a maximum observed error of $0.09$, which is acceptable given the measurement noise.
\footnote{The verification of each limb of STA will be presented in Section C of Appendix.}

\subsection{Loss Function}

Our training objective consists of three main components that correspond to distinct stages in our generation pipeline: geometry-guided denoising diffusion, and control mask prediction.

\paragraph{Geometry-guided diffusion loss.}
Following SGDiffusion~\cite{desobav2}, we supervise the denoising network using a geometry-aware conditional control feature extracted by KPLM and STA. These modules provide spatial priors that are injected via ControlNet. The diffusion noise prediction loss is formulated as:

\begin{equation}
\mathcal{L}_{\text{mwsg}} \!=\! \mathrm{E}_{t,\epsilon \sim \mathcal{N}(0,1)} \left[ \left\| W_{fs} \circ \left( \epsilon - \hat{\epsilon}_\psi(z_t, t, c_{\text{geo}}) \right) \right\|_2^2 \right],
\label{EQ8}
\end{equation}
% \begin{equation}
% \mathcal{L}_{\text{mwsg}} = \mathrm{E}_{t,\epsilon \sim \mathcal{N}(0,1)} \left[ \left\| W_{fs} \circ \left( \epsilon - \hat{\epsilon}_\psi(z_t, t, c_{\text{geo}}) \right) \right\|_2^2 \right],
% \label{EQ8}
% \end{equation}
where $c_{\text{geo}}$ denotes the concatenated geometric condition derived from KPLM and STA, $\hat{\epsilon}_\psi$ is the noise predicted by the denoising network at time step $t$, $W_{fs}$ is the soft foreground shadow mask, highlighting training focus on shadow regions.

\paragraph{Foreground shadow mask loss.}

\begin{figure*}[!ht]
\centering  %图片全局居中
\includegraphics[width=17cm,height=10cm]{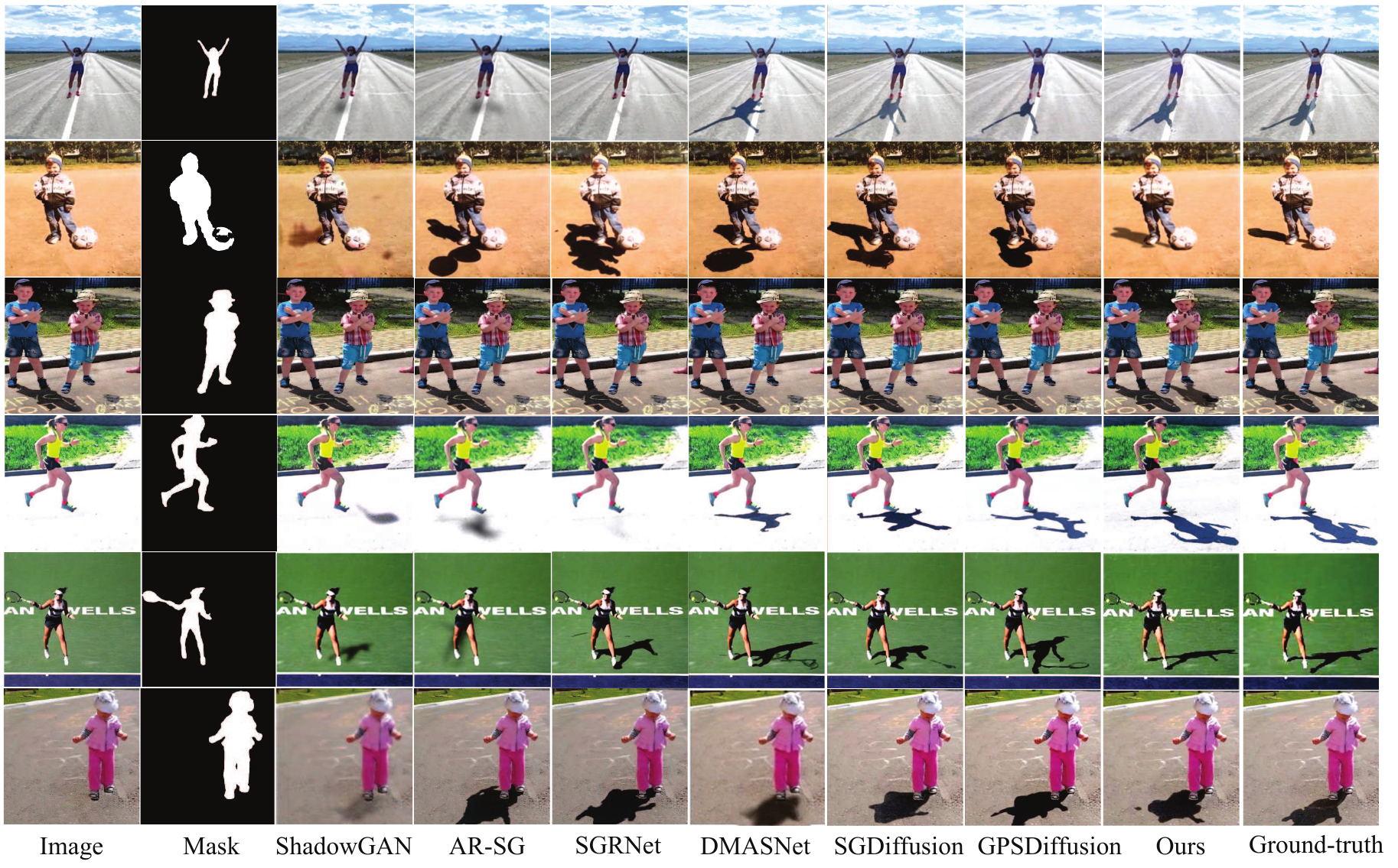}
\caption{ Visual comparison with state-of-the-art methods on DESOBAv2 dataset. From left to right, we show composite image, foreground
object mask, results of ShadowGAN\cite{53shadowgan}, AR-SG\cite{ar-sg}, SGRNet\cite{desoba}, DMASNet\cite{dmasnet}, SGDiffusion\cite{desobav2}, GPSDiffusion\cite{gpsdiff}, Ours, Ground-truth.}
\label{desobav2t}
\end{figure*}

To ensure that the generated shadows are correctly located with respect to the foreground object, we supervise the predicted mask $\hat{M}_{fs}$ with an $L_1$ loss against the ground-truth $M_{fs}$:

\begin{equation}
\mathcal{L}_{\text{mask}} = \left\| \hat{M}_{fs} - M_{fs} \right\|_1.
\label{EQ9}
\end{equation}

Optionally, dice loss or binary cross-entropy can be added to stabilize training, but we use $L_1$ loss by default.
Finally, we summarize the mask prediction loss in Eq.~\ref{EQ8} and
weighted noise loss in Eq.~\ref{EQ9} as $\mathcal{L}_{all}=\mathcal{L}_{mswg}+\lambda\mathcal{L}_{mask}$,
where $\lambda$ is a trade-off parameter.

\section{Experiments}
\subsection{Experiments Setting}
For KPLM, we need to obtain the key point coordinates of each image and store them in json form. Each json includes information such as image number, pose: front, side, key point coordinates, etc. 
For STA, we need Ground Truth as the guide, and we need to obtain the coordinates of points A, B, and C for each Ground Truth. 
Following~\cite{gpsdiff}, we divide DESOBAv2 into 21,088 training images with 27,718 tuples (composite image, foreground object mask,
ground-truth image) and 487 test images with 855 tuples, including both BOS (with background object-shadow pairs) and BOS-free images in the test set. Our Method is implemented with PyTorch~\cite{pytorch}. We use AdamW optimizer~\cite{adam} with a fixed learning rate of 0.0001. All experiments are conducted on 4 NVIDIA RTX 4090 GPUs. 

\begin{table*}[!ht]
\centering
\label{desobav2}
\begin{tabular}{lcccccccccccc}
\toprule
\multirow{2}{*}{Method} & \multicolumn{6}{c}{BOS Test Image} & \multicolumn{6}{c}{BOS-free Test Image} \\ % 合并6列
\cmidrule(lr){2-7}\cmidrule(lr){8-13} % 部分横线，仅覆盖第2-7列% 部分横线，仅覆盖第2-7列
 & GR $\downarrow$     & LR $\downarrow$     & GS $\uparrow$     & LS $\uparrow$     & GB $\downarrow$    & LB $\downarrow$ 
 & GR $\downarrow$    & LR $\downarrow$    & GS $\uparrow$    & LS $\uparrow$    & GB $\downarrow$    & LB $\downarrow$ \\ % 指标名称
\midrule
ShadowGAN    & 7.511  & 67.464 & 0.961  & 0.197  & 0.446  & 0.890  & 17.325 & 76.508 & 0.901 & 0.060 & 0.425 & 0.842 \\
Mask-SG     & 8.997  & 79.418 & 0.951  & 0.180  & 0.500  & 1.000  & 19.338 & 94.327 & 0.906 & 0.044 & 0.500 & 1.000\\
AR-SG       & 7.335  & 58.037 & 0.961  & 0.241  & 0.383  & 0.761  & 16.067 & 63.713 & 0.908 & 0.104 & 0.349 & 0.682 \\
SGRNet      & 7.184  & 68.255 & 0.964  & 0.206  & 0.301  & 0.596  & 15.596 & 60.350 & 0.909 & 0.100 & 0.271 & 0.534 \\
DMASNet      & 8.256  & 59.380 & 0.961  & 0.228  & 0.276  & 0.547  & 18.725 & 86.694 & 0.913 & 0.055 & 0.297 & 0.574\\
SGDiffusion  & 6.098  & 53.611 & 0.971  & 0.370  & 0.245  & 0.487  & 15.110 &55.874 & 0.913 & 0.117 & 0.233 & 0.452\\
GPSDiffusion & 5.896  & 46.713 & 0.966  & 0.374  & 0.213  & 0.423  & 13.809 & 55.616 & 0.917 & 0.166 & \textbf{0.197} & 0.384 \\
\textbf{Ours}         & \textbf{4.486} & \textbf{46.190} & \textbf{0.972} & \textbf{0.376} & \textbf{0.205} & \textbf{0.418} & \textbf{12.620} & \textbf{54.243} & \textbf{0.941} & \textbf{0.247} & 0.198 & \textbf{0.382} \\
\bottomrule
\end{tabular}
\caption{ The results of different methods on DESOBAv2 dataset. The best results are highlighted in boldface.\textbf{GSSIM, LSSIM, and LBer} are indicators that can measure appearance realism, while \textbf{GRMSE, LRMSE} and \textbf{GBer} are indicators that can measure geometric accuracy.
}
\label{desobav2-t}
\end{table*}

\begin{table*}[!ht]
\centering
\label{bos}
\begin{tabular}{lcccccccccc}
\toprule
\multirow{1}{*}{Method} & \multicolumn{4}{c}{BOS Test Image}
&
\multicolumn{4}{c}{BOS-free Test Image}
\\ % 合并6列
\midrule % 部分横线，仅覆盖第2-7列
Evaluation matric & GRMSE $\downarrow$    & LRMSE $\downarrow$    & GSSIM $\uparrow$    & LSSIM* $\uparrow$ & GRMSE $\downarrow$    & LRMSE  $\downarrow$    & GSSIM $\uparrow$    & LSSIM* $\uparrow$ \\ % 指标名称
\midrule
Pix2Pix   & 7.659  & 75.346 & 0.927  & 0.249 & 18.875  & 81.444 & 0.856  & 0.110   \\
Pix2Pix-Res      & 18.305  & 81.966 & 0.901  & 0.107  & 5.961  & 76.046 & 0.971  & 0.253  \\
ShadowGAN        & 5.985  & 78.413 & 0.986  & 0.240  & 19.306  & 87.017 & 0.918  & 0.078 \\
Mask-ShadowGAN       & 8.287  & 79.212 & 0.953  & 0.245  & 19.475  & 83.457 & 0.891  & 0.109 \\
ARShadowGAN      & 6.481  & 75.099 & 0.983  & 0.251  & 18.723  & 81.272 & 0.917  & 0.109 \\
SGRNet  & 4.754  & 61.762 & 0.988  & 0.380 & 15.128  & 61.439 & 0.927  & 0.183 \\
GPSDiffusion &  3.613 & 39.843 & 0.991  & 0.415  &  12.603 & 60.312 & 0.931  & 0.197 \\
\textbf{Ours}         & \textbf{3.012} & \textbf{38.720} & \textbf{0.997} & \textbf{0.423} & \textbf{11.974} & \textbf{59.108} & \textbf{0.933} & \textbf{0.198} \\
\bottomrule
\end{tabular}
\caption{ The results of different methods on DESOBA dataset. The best results are highlighted in boldface.\textbf{GSSIM, LSSIM} are indicators that can measure appearance realism, while \textbf{GRMSE, LRMSE} are indicators that can measure geometric accuracy.}
\label{desoba}
\end{table*}
\subsection{Experiments with IC-Light}

\begin{figure}[!ht]
\centering  %图片全局居中
\includegraphics[width=9cm,height =5.4cm]{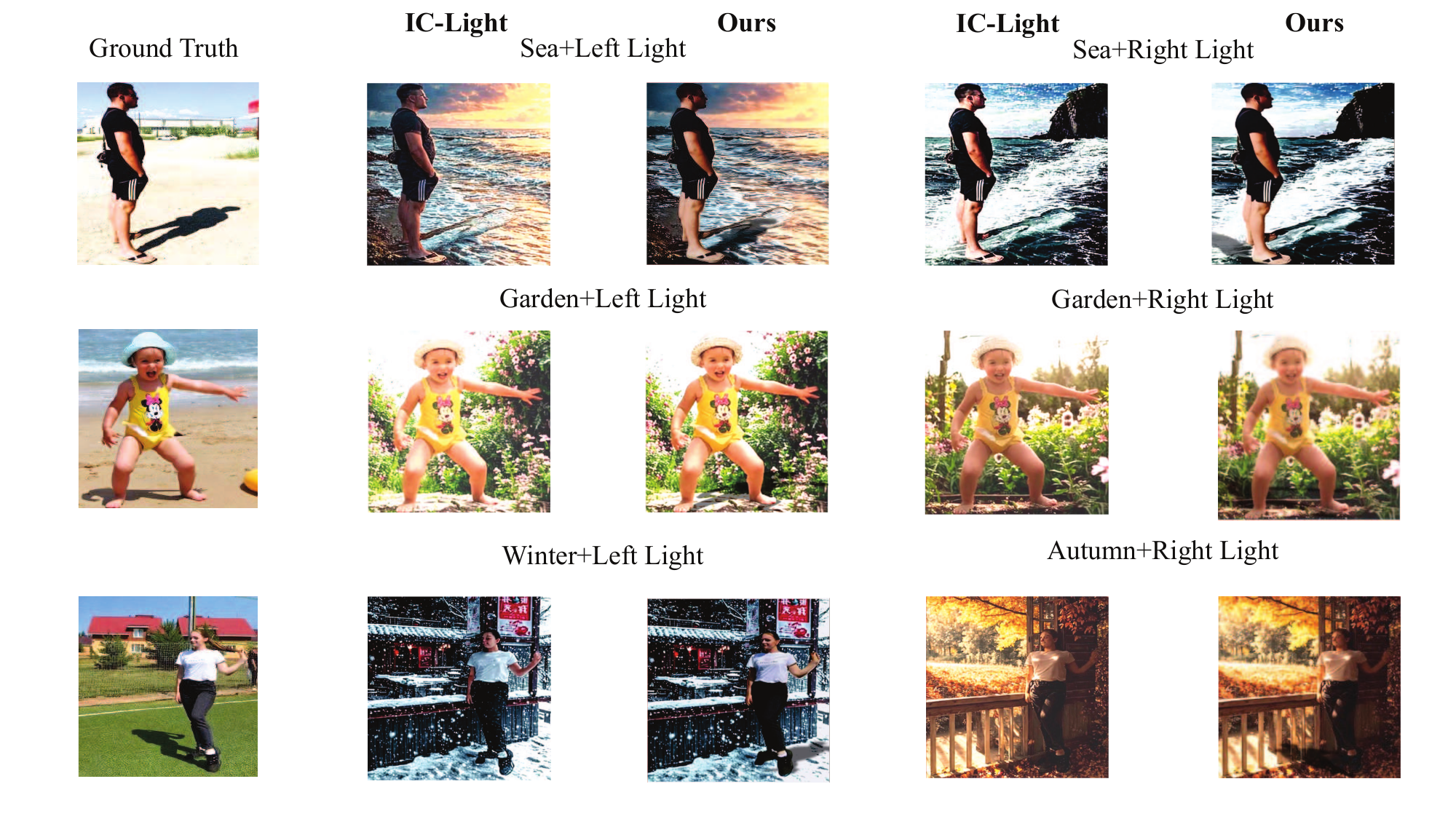}
\caption{Visualization results of shadow generation for IC-Light relighting images by our method.}
\label{iclight-e}
\end{figure}

\begin{figure}[!ht]
    \centering
    \includegraphics[width=8cm]{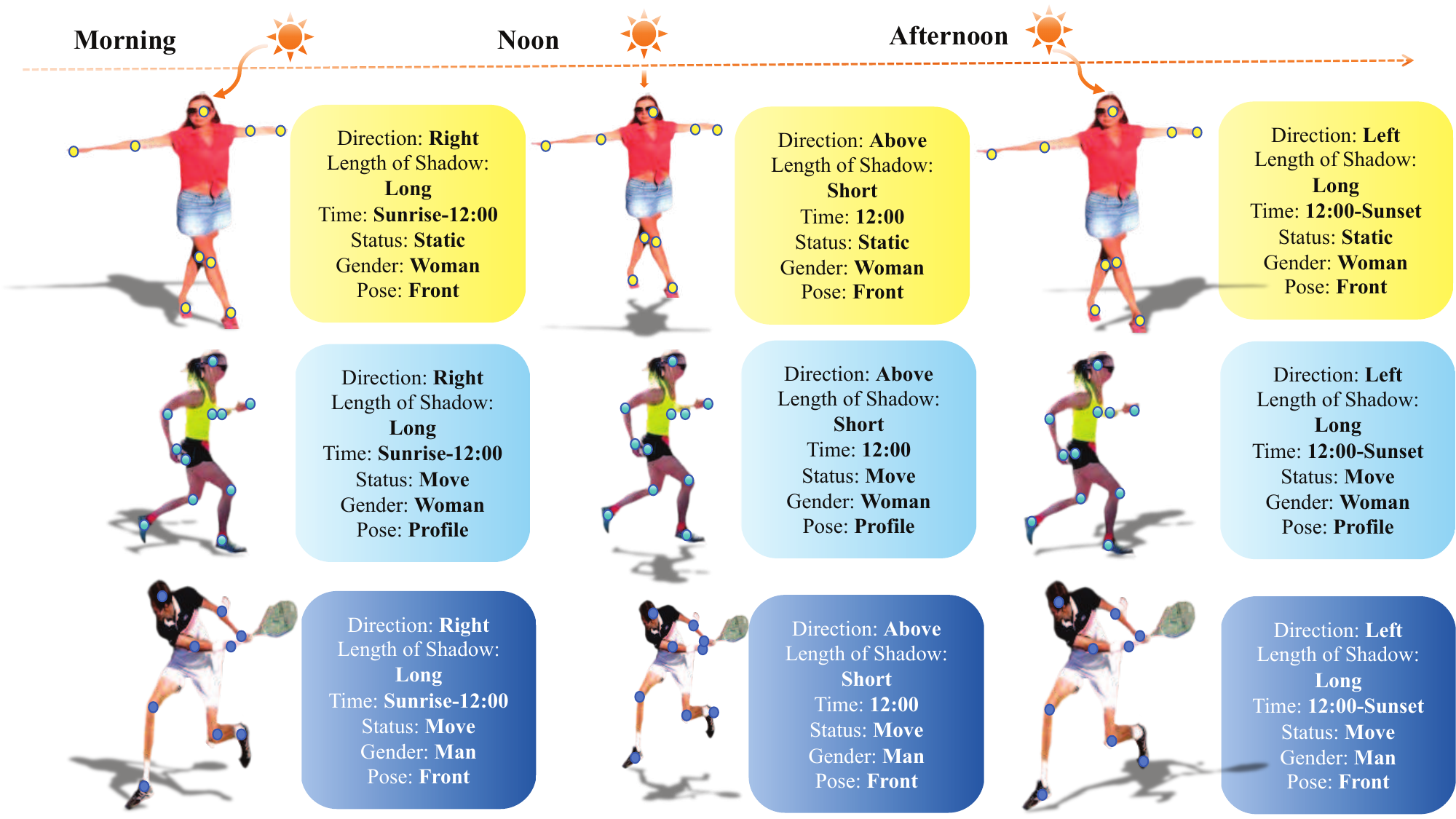}
    \caption{KPLM with Different Light.}
    \label{diff-light}
\end{figure}

The goal of this paper is to generate physically accurate shadows for images relit by IC-Light. We deploy the open-source IC-Light model on an NVIDIA RTX 4090, keeping the output resolution consistent with that of the input image. To ensure generalizability, we evaluate under varying scene types, seasonal conditions, and lighting directions.
Although IC-Light excels in relighting appearance, it does not produce realistic ground shadows. Our method, when applied to the relit outputs, can synthesize plausible and geometry consistent shadows. For quantitative evaluation, we use ground-truth images as reference and compare PSNR, SSIM, and LPIPS between IC-Light and our method, as shown in Table~\ref{t3}. IC-Light's scores on these metrics are naturally lower due to the lack of shadow data post-relighting, but this is to be expected and does not reflect its original relighting performance.
Figure~\ref{iclight-e} visualizes the added shadows by our method. Since our pipeline is conditioned on both the relit image and the original structure, the generated shadows align well with the actual scene semantics. 
As demonstrated in Figure~\ref{diff-light}, our KPLM module remains effective under diverse lighting directions.
\footnote{More details on IC-Light will be presented in Section A of Appendix.}

\begin{table}[!ht]
\centering
\begin{tabular}{ccccc}
\hline
Prompt & Method & PSNR $\uparrow$ & SSIM $\uparrow$ & LPIPS $\downarrow$ \\
\hline
\multirow{2}{*}{Different Scene} 
  & IC-Light & 27.961 & 0.166 & 0.612 \\
  & \textbf{Ours} & \textbf{28.140} & \textbf{0.235} & \textbf{0.590} \\
\hline
\multirow{2}{*}{Different Light} 
  & IC-Light & 27.869 & 0.182 & 0.582 \\
  & \textbf{Ours} & \textbf{27.972} & \textbf{0.207} & \textbf{0.563} \\
\hline
\multirow{2}{*}{Different Season} 
  & IC-Light & 27.759 & 0.200 & 0.584 \\
  & \textbf{Ours} & \textbf{27.767} & \textbf{0.249} & \textbf{0.558} \\
\hline
\end{tabular}
\caption{Comparison under different prompts on DESOBAv2.}
\label{t3}
\end{table}

\subsection{Experiments on DESOBA}
The results of the quantitative comparison are summarized in Table~\ref{desoba}. In contrast to DESOBAv2, our comparisons include Pix2Pix~\cite{pix2pix} and its residual variant, Pix2Pix-Res. Our method consistently outperforms all baseline approaches across all metrics. Although the DESOBA dataset is smaller in scale compared to DESOBAv2, our approach still achieves state-of-the-art performance. Specifically, our method yields the lowest GRMSE and LRMSE, as well as the highest GSSIM and LSSIM* scores, clearly demonstrating its effectiveness. In fairness, all methods were evaluated under the same pre-processing protocol, and the visualization results for DESOBA are included in the supplementary materials.
\footnote{Visualization results on DESOBA are included in the Appendix.}

\subsection{Experiments on DESOBAv2}

As shown in Table~\ref{desobav2-t}, our method achieves the best performance across most key metrics, including GRMSE, GSSIM, LSSIM, and LB, demonstrating a clear overall advantage. Although our GBer score is slightly higher than the GPSdiffusion~\cite{gpsdiff}, this marginal difference of 0.001 falls within the range of experimental variation and has negligible impact on overall performance. This minor increase in boundary error may result from our model’s design, which emphasizes structural consistency and semantic alignment with human posture. Such a design choice may lead to slightly softer transitions at certain boundary regions. Nevertheless, our superior LB score indicates that our method still maintains accurate delineation in critical boundary areas. Therefore, we believe our method strikes a better balance between structural precision and visual realism.
\footnote{Real composite image visualization on DESOBAv2 will be presented in the Appendix.}
\subsection{Ablation}
We conduct ablation studies on BOS test
images to investigate the impact of each module using four
metrics: GR, LR, GB, and LB.
Firstly, we trained the base model of ControlNet, which had no geometry before, as shown in the first row of Table~\ref{ablation}. Secondly, we injected STA into ControlNet to explore the effect. The result in the second line confirmed that the shadow triangle algorithm is useful. Compare the third line and the first line to verify the validity of KPLM. As mentioned in Line 5, our mature model achieves the best performance. To further demonstrate our effectiveness, we evaluated the metrics on the DESOBAv2 test set.
\begin{table}[!ht]
\centering
\scalebox{0.8}{
\begin{tabular}{cccccccc}
\hline
    Row & Base & KPLM & STA & GR $\downarrow$ & LR $\downarrow$ & GB $\downarrow$ & LB $\downarrow$ \\
\hline
    1 & $+$ & $-$ & $-$ & 14.317 & 61.735 & 0.214 & 0.441  \\ 
    2 & $+$ & $-$ & $+$ & 13.172 & 55.384 & 0.201 & 0.397 \\
    3 & $+$ & $+$ & $-$ & 13.213 & 55.744 & 0.207 & 0.391 \\
    4 & $+$ & $+$ & $+$ & 12.620 & 54.243 & 0.198 & 0.382 \\
\hline
\end{tabular}
}
\caption{\textbf{Ablation studies of our method on BOS test images
from DESOBAv2 dataset.} “Base” means ControlNet base model.}
% “KPLM” means using Keypoints Linear Model to get key points. “STA” means using the Shadow Triangle Algorithm to determine the position of the shadow.}
\label{ablation}
\end{table}

\section{Conclusion}
In this paper, our framework for generating shadows after IC-Light relighting is presented. The Keypoints Linear Model (KPLM) and Shadow Triangle Algorithm (STA) are combined to identify the spatial coordinates of keypoints that correspond to shadows induced by the human body. Both modules obtain all necessary geometric information during preprocessing and inject it directly into the control encoder. Then performed using a diffusion-based model to generate shadow. In future work, we will explore more lightweight diffusion models, aiming to maintain generation quality while significantly reducing model complexity. Additionally, we will investigate further structural refinements to enhance both the efficiency and fidelity of the shadow.

\bibliography{aaai2026}

@inproceedings{
    iclight,
    title={Scaling In-the-Wild Training for Diffusion-based Illumination Harmonization and Editing by Imposing Consistent Light Transport},
    author={Lvmin Zhang and Anyi Rao and Maneesh Agrawala},
    booktitle={The Thirteenth International Conference on Learning Representations},
    year={2025},
    url={https://openreview.net/forum?id=u1cQYxRI1H}
}

@article{desoba,
  title={Shadow Generation for Composite Image in Real-world Scenes},
  author={Hong, Yan and Niu, Li and Zhang, Jianfu},
  journal={AAAI},
  year={2022}
}

@inproceedings{desobav2,
  title={Shadow generation for composite image using diffusion model},
  author={Liu, Qingyang and You, Junqi and Wang, Jianting and Tao, Xinhao and Zhang, Bo and Niu, Li},
  booktitle={Proceedings of the IEEE/CVF Conference on Computer Vision and Pattern Recognition},
  pages={8121--8130},
  year={2024}
}

@inproceedings{1,
  title={Realtime estimation of illumination direction for augmented reality on mobile devices},
  author={Arief, Ibrahim and McCallum, Simon and Hardeberg, Jon Yngve},
  booktitle={Color and Imaging Conference},
  volume={20},
  pages={111--116},
  year={2012},
  organization={Society of Imaging Science and Technology}
}

@inproceedings{7,
  title={Deep parametric indoor lighting estimation},
  author={Gardner, Marc-Andr{\'e} and Hold-Geoffroy, Yannick and Sunkavalli, Kalyan and Gagn{\'e}, Christian and Lalonde, Jean-Fran{\c{c}}ois},
  booktitle={Proceedings of the IEEE/CVF International Conference on Computer Vision},
  pages={7175--7183},
  year={2019}
}

@article{53shadowgan,
  title={Shadowgan: Shadow synthesis for virtual objects with conditional adversarial networks},
  author={Zhang, Shuyang and Liang, Runze and Wang, Miao},
  journal={Computational Visual Media},
  volume={5},
  pages={105--115},
  year={2019},
  publisher={Springer}
}

@inproceedings{mask-sg,        
  title={{Mask-ShadowGAN}: Learning to Remove Shadows from Unpaired Data},         
  author={Hu, Xiaowei and Jiang, Yitong and Fu, Chi-Wing and Heng, Pheng-Ann},         
  booktitle={Proceedings of the IEEE/CVF International Conference on Computer Vision},       
  year={2019}
}

@InProceedings{ar-sg,  
  title = {ARShadowGAN: Shadow Generative Adversarial Network for Augmented Reality in Single Light Scenes},
  author = {Liu, Daquan and Long, Chengjiang and Zhang, Hongpan and Yu, Hanning and Dong, Xinzhi and Xiao, Chunxia},
  booktitle = {The IEEE Conference on Computer Vision and Pattern Recognition},
  month = {June},
  year = {2020}
}

@misc{dmasnet,
      title={Shadow Generation with Decomposed Mask Prediction and Attentive Shadow Filling}, 
      author={Xinhao Tao and Junyan Cao and Yan Hong and Li Niu},
      year={2024},
      eprint={2306.17358},
      archivePrefix={arXiv},
      primaryClass={cs.CV},
      url={https://arxiv.org/abs/2306.17358}, 
}

@inproceedings{gpsdiff,
  title={Shadow Generation Using Diffusion Model with Geometry Prior},
  author={Zhao, Haonan and Liu, Qingyang and Tao, Xinhao and Niu, Li and Zhai, Guangtao},
  booktitle={Proceedings of the Computer Vision and Pattern Recognition Conference},
  pages={7603--7612},
  year={2025}
}

@inproceedings{pix2pix,
  title={Image-to-image translation with conditional adversarial networks},
  author={Isola, Phillip and Zhu, Jun-Yan and Zhou, Tinghui and Efros, Alexei A},
  booktitle={Proceedings of the IEEE conference on computer vision and pattern recognition},
  pages={1125--1134},
  year={2017}
}

@article{openpose,
  author = {Z. {Cao} and G. {Hidalgo Martinez} and T. {Simon} and S. {Wei} and Y. A. {Sheikh}},
  journal = {IEEE Transactions on Pattern Analysis and Machine Intelligence},
  title = {OpenPose: Realtime Multi-Person 2D Pose Estimation using Part Affinity Fields},
  year = {2019}
}

@inproceedings{openposehand,
  author = {Tomas Simon and Hanbyul Joo and Iain Matthews and Yaser Sheikh},
  booktitle = {Proceedings of the IEEE/CVF Conference on Computer Vision and Pattern Recognition},
  title = {Hand Keypoint Detection in Single Images using Multiview Bootstrapping},
  year = {2017}
}

@inproceedings{openposerealtime,
  author = {Zhe Cao and Tomas Simon and Shih-En Wei and Yaser Sheikh},
  booktitle = {Proceedings of the IEEE/CVF Conference on Computer Vision and Pattern Recognition},
  title = {Realtime Multi-Person 2D Pose Estimation using Part Affinity Fields},
  year = {2017}
}

@inproceedings{openposecpm,
  author = {Shih-En Wei and Varun Ramakrishna and Takeo Kanade and Yaser Sheikh},
  booktitle = {Proceedings of the IEEE/CVF Conference on Computer Vision and Pattern Recognition},
  title = {Convolutional pose machines},
  year = {2016}
}

@incollection{smpl,
  title={SMPL: A skinned multi-person linear model},
  author={Loper, Matthew and Mahmood, Naureen and Romero, Javier and Pons-Moll, Gerard and Black, Michael J},
  booktitle={Seminal Graphics Papers: Pushing the Boundaries, Volume 2},
  pages={851--866},
  year={2023}
}

@inproceedings{eccv,
  title={Controllable shadow generation using pixel height maps},
  author={Sheng, Yichen and Liu, Yifan and Zhang, Jianming and Yin, Wei and Oztireli, A Cengiz and Zhang, He and Lin, Zhe and Shechtman, Eli and Benes, Bedrich},
  booktitle={European Conference on Computer Vision},
  pages={240--256},
  year={2022},
  organization={Springer}
}

@misc{mediapipe,
  author = {Google},
  title = {MediaPipe: A Framework for Building Perception Pipelines},
  year = {2020},
  howpublished = {\url{https://mediapipe.dev}},
  note = {Accessed: 2025-07-11}
}

@article{composite,
  title={Making images real again: A comprehensive survey on deep image composition},
  author={Niu, Li and Cong, Wenyan and Liu, Liu and Hong, Yan and Zhang, Bo and Liang, Jing and Zhang, Liqing},
  journal={arXiv preprint arXiv:2106.14490},
  year={2021}
}

@article{intro_1,
  title={Satellite images analysis for shadow detection and building height estimation},
  author={Liasis, Gregoris and Stavrou, Stavros},
  journal={ISPRS Journal of Photogrammetry and Remote Sensing},
  volume={119},
  pages={437--450},
  year={2016},
  publisher={Elsevier}
}

@article{intro_2,
  title={A shadow-overlapping algorithm for estimating building heights from VHR satellite images},
  author={Kadhim, Nada and Mourshed, Monjur},
  journal={IEEE Geoscience and remote sensing letters},
  volume={15},
  number={1},
  pages={8--12},
  year={2017},
  publisher={IEEE}
}

@article{intro_3,
  title={Synthetic datasets for autonomous driving: A survey},
  author={Song, Zhihang and He, Zimin and Li, Xingyu and Ma, Qiming and Ming, Ruibo and Mao, Zhiqi and Pei, Huaxin and Peng, Lihui and Hu, Jianming and Yao, Danya and others},
  journal={IEEE Transactions on Intelligent Vehicles},
  volume={9},
  number={1},
  pages={1847--1864},
  year={2023},
  publisher={IEEE}
}

@article{ddsm,
  title={Denoising diffusion step-aware models},
  author={Yang, Shuai and Chen, Yukang and Wang, Luozhou and Liu, Shu and Chen, Yingcong},
  journal={arXiv preprint arXiv:2310.03337},
  year={2023}
}

@inproceedings{controlnet,
  title={Adding conditional control to text-to-image diffusion models},
  author={Zhang, Lvmin and Rao, Anyi and Agrawala, Maneesh},
  booktitle={Proceedings of the IEEE/CVF international conference on computer vision},
  pages={3836--3847},
  year={2023}
}

@inproceedings{ldm,
  title={High-Resolution Image Synthesis with Latent Diffusion Models},
  author={Rombach, Robin and Blattmann, Andreas and Lorenz, Dominik and Esser, Patrick and Ommer, Bj{\"o}rn},
  booktitle={Proceedings of the IEEE/CVF Conference on Computer Vision and Pattern Recognition},
  pages={10684--10695},
  year={2022},
  note={arXiv preprint arXiv:2112.10752 [cs.CV]},
  url={https://arxiv.org/abs/2112.10752}
}

@inproceedings{lsgan,
  title={Least Squares Generative Adversarial Networks},
  author={Mao, Xudong and Li, Qing and Xie, Haoran and Lau, Raymond Y. K. and Wang, Zhen and Smolley, Stephen Paul},
  booktitle={Proceedings of the IEEE/CVF International Conference on Computer Vision},
  pages={2813--2821},
  year={2017},
  note={arXiv preprint arXiv:1611.04076 [cs.CV]},
  url={https://arxiv.org/abs/1611.04076}
}

@inproceedings{gan,
  title={Generative Adversarial Networks},
  author={Ian J. Goodfellow and Jean Pouget-Abadie and Mehdi Mirza and Bing Xu and David Warde-Farley and Sherjil Ozair and Aaron Courville and Yoshua Bengio},
  booktitle={Advances in Neural Information Processing Systems (NeurIPS)},
  volume={27},
  pages={2672--2680},
  year={2014},
}

@inproceedings{patchgan,
  title={Image-to-Image Translation with Conditional Adversarial Networks},
  author={Isola, Phillip and Zhu, Jun-Yan and Zhou, Tinghui and Efros, Alexei A},
  booktitle={Proceedings of the IEEE/CVF Conference on Computer Vision and Pattern Recognition},
  pages={1125--1134},
  year={2017},
  url={https://arxiv.org/abs/1611.07004}
}

@article{pytorch,
  title={Pytorch: An imperative style, high-performance deep learning library},
  author={Paszke, Adam and Gross, Sam and Massa, Francisco and Lerer, Adam and Bradbury, James and Chanan, Gregory and Killeen, Trevor and Lin, Zeming and Gimelshein, Natalia and Antiga, Luca and others},
  journal={Advances in neural information processing systems},
  volume={32},
  year={2019}
}

@article{adam,
  title={Decoupled weight decay regularization},
  author={Loshchilov, Ilya and Hutter, Frank},
  journal={arXiv preprint arXiv:1711.05101},
  year={2017}
}

@article{pose2light,
  title={Pose2Light: Learning to Estimate Illumination from Human Pose},
  author={Liu, Dong and Qiu, Qi and Shen, Ying},
  journal={IEEE Transactions on Circuits and Systems for Video Technology},
  year={2022},
  volume={32},
  number={4},
  pages={2135--2148}
}

\end{document}

\appendix

\section{A: More details of limitation on IC-Light}

\begin{figure}[htbp]
    \centering
    \includegraphics[width=8cm,height=4.5cm]{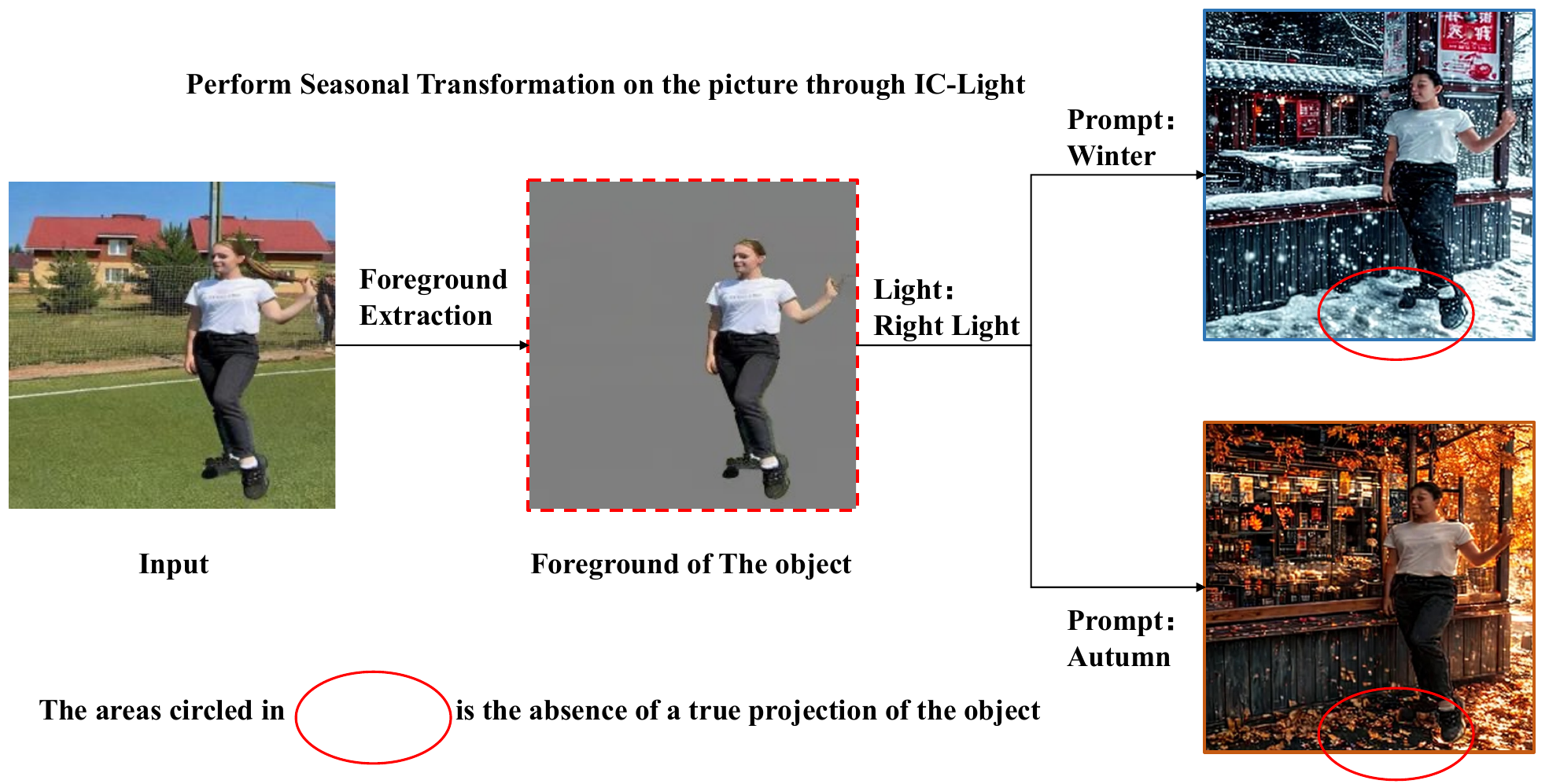}
    \caption{Instances of missing shadows in IC-Light.}
    \label{iclight}
\end{figure}
As shown in the Figure~\ref{iclight}, it is demonstrated that IC-Light~\cite{iclight} handles the Relighting details very well (clothing fold light details, facial details, etc.), but the projection on the real ground causes the deficiency. To further investigate this limitation, we conducted additional comparisons using varied prompts, lighting conditions, and alternative visualizations of both shadow and shadow-free images.
\begin{figure}[htbp]
    \centering
    \includegraphics[width=7cm]{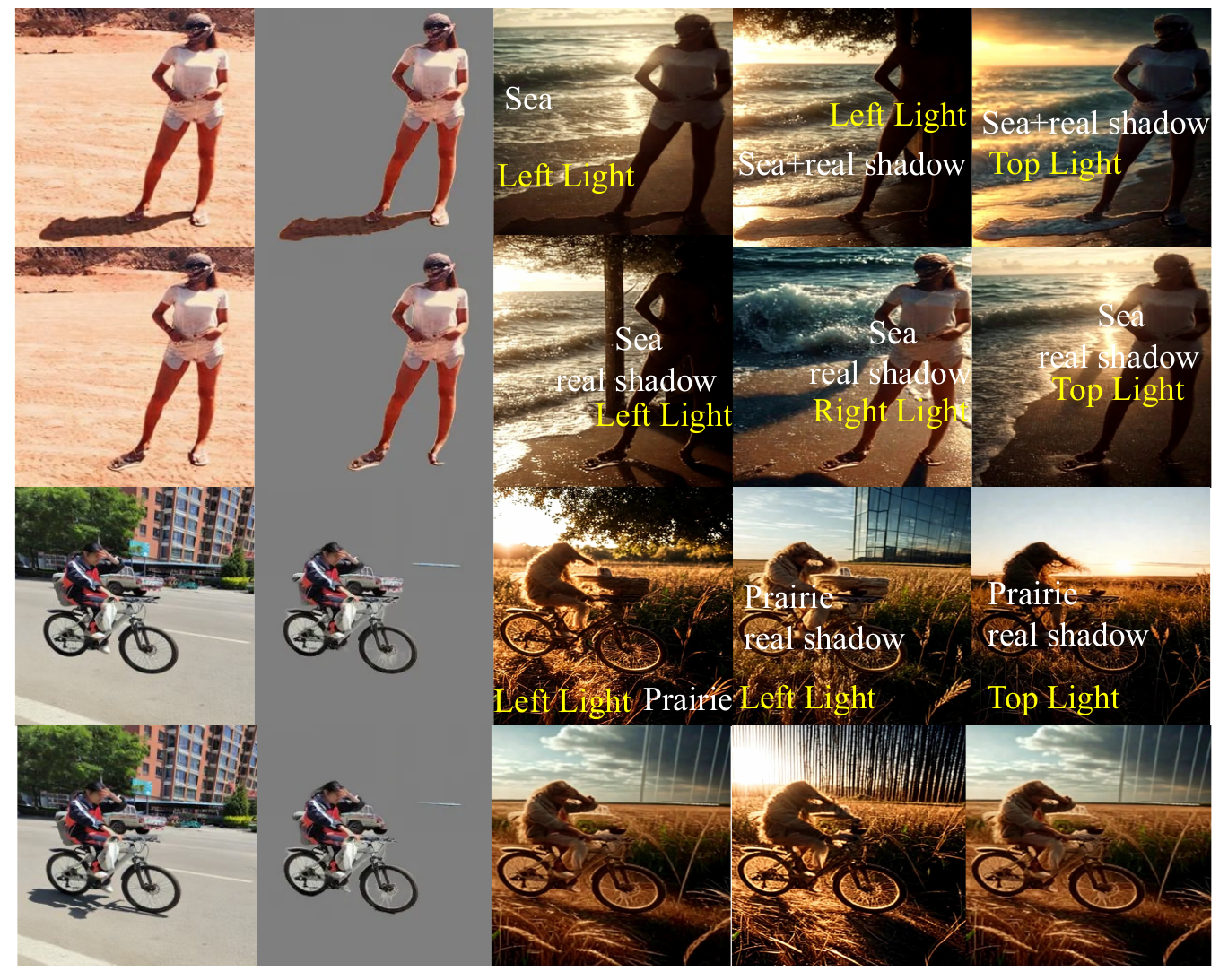}
    \caption{Experiment on IC-Light.}
    \label{iclight-ex1}
\end{figure}

\begin{figure}[!ht]
    \centering
    \includegraphics[width=7cm]
    {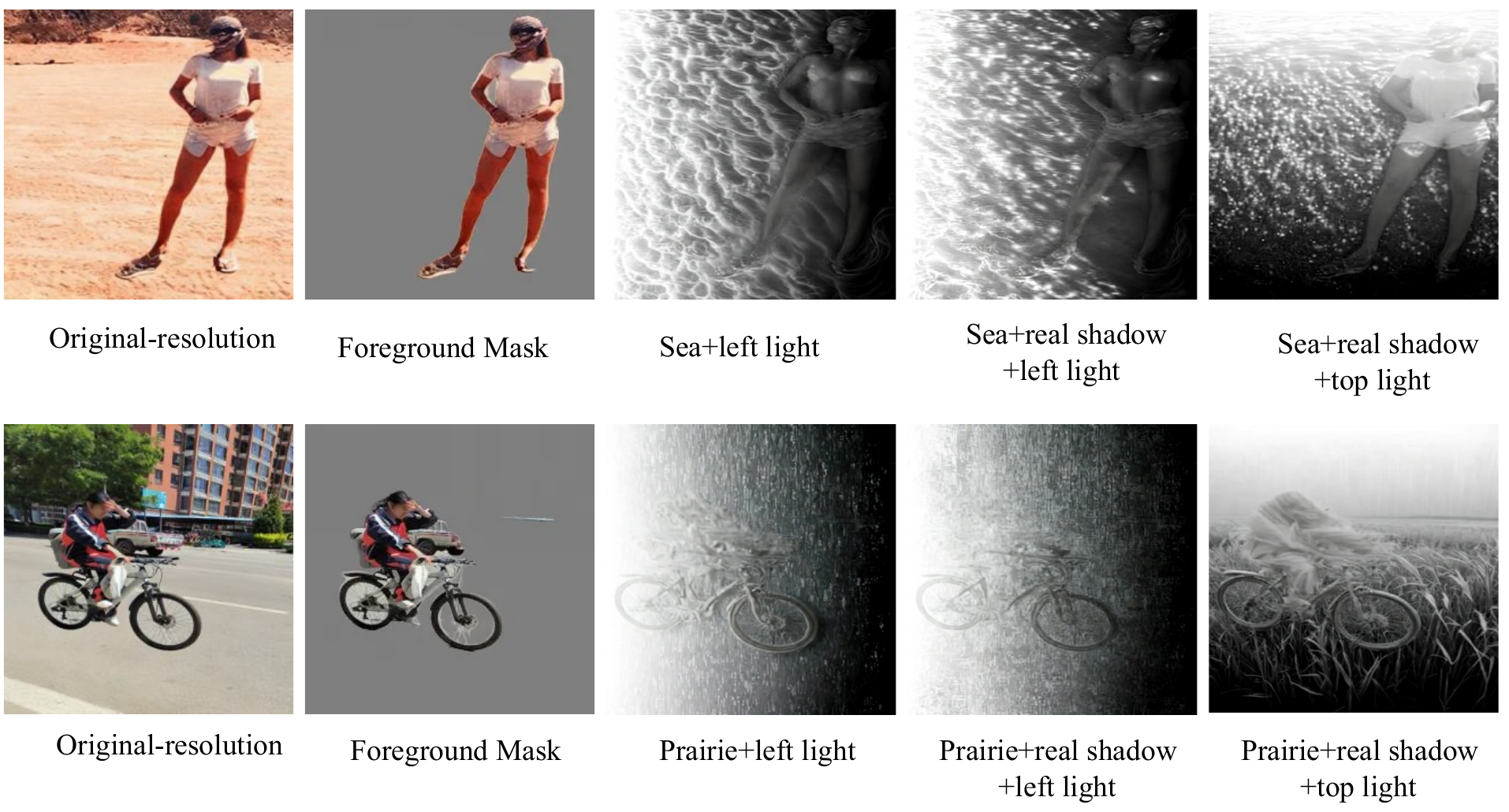}
    \caption{Different parm settings on IC-Light.}
    \label{iclight-ex3}
\end{figure}

\begin{figure}[!ht]
    \centering
    \includegraphics[width=7cm]
    {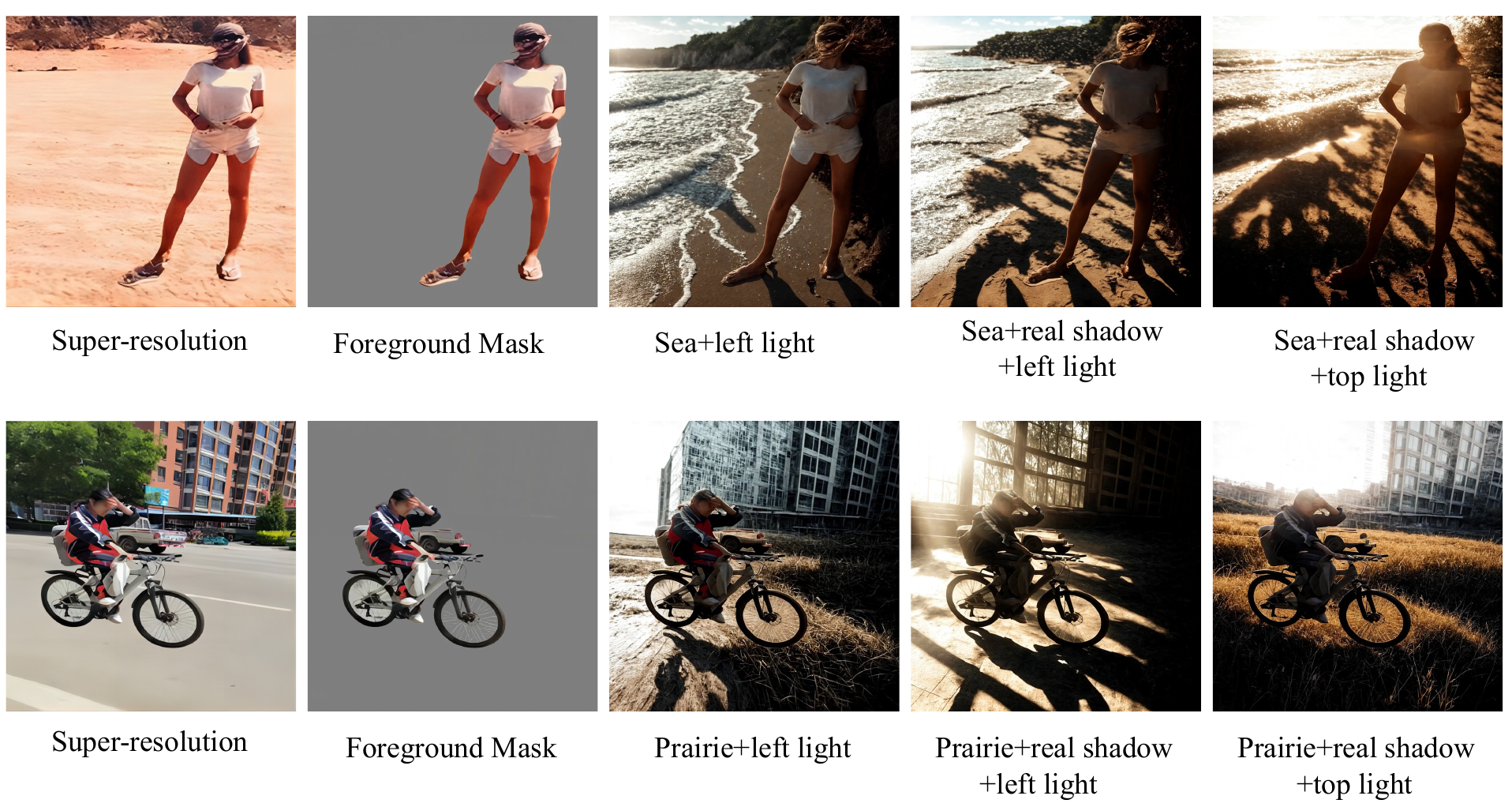}
    \caption{Different resolution ratio on IC-Light.}
    \label{iclight-ex2}
\end{figure}

\begin{table*}[!ht]
\centering
\scalebox{0.85}{\begin{tabular}{c|c|c|c|c|c|c|c|c|c}
%{p{1.5cm}|p{1.3cm}|p{1.8cm}|p{1.6cm}}
\hline
    Parameters & Setting & Image & Resolution Ratio & Setting & Image & Resolution Ratio & Setting & Image & Resolution Ratio \\
\hline
    Steps & 25 & 2 & 256*256 & 25 & 3 & 256*256 & 25 & 4 & 1024*1024 \\
\hline
    CFG & 2 & 2 & 256*256 & 3 & 3 & 256*256 & 2 & 4 & 1024*1024 \\
\hline
    Denoise & 0.9 & 2 & 256*256 & 0.5 & 3 & 256*256 & 0.9 & 4 & 1024*1024  \\ 
\hline
    Highres Scale & 1.5 & 2 & 256*256 & 1.5 & 3 & 256*256 & 1.5 & 4 & 1024*1024 \\
\hline
    Highres Denoise & 0.5 & 2 & 256*256 & 0.5 & 3 & 256*256 & 0.5 & 4 & 1024*1024 \\
\hline
\end{tabular}
}
\caption{Evaluation with different parameters on IC-Light.}
\label{iclight-ex}
\end{table*}

\section{B: More details of mask experiment on KPLM}
In the main text, we mentioned that KPLM adopts a strategy of extracting key points with $9$ key points and $1$ key block. In the experimental section, we also conducted ablation experiments to verify the effectiveness of KPLM and STA. In this section, we will verify why this strategy is effective. We will compare the three different representation methods of $9$ key points, $14$ key points and $9$ key points in $1$ block to prove that the key point extraction of $9$ key points in $1$ key block is the most effective.
\begin{table}[htbp]
    \centering
    \begin{tabular}{c|c|c|c}
    \hline
        Method & IoU $\uparrow$ & Time (ms) $\downarrow$ & Numbers $\downarrow$ \\
    \hline
        $9$ key points & 0.402 & \textbf{5210} & \textbf{9} \\
    \hline
        $14$ key points & 0.477 & 5503 & 14 \\
    \hline
        KPLM & \textbf{0.535} & 5403 & 14 \\
    \hline
    \end{tabular}
    \caption{Different points settings on human body.}
    \label{kplm}
\end{table}

\begin{figure}[htbp]
    \centering
    \includegraphics[width=6cm]{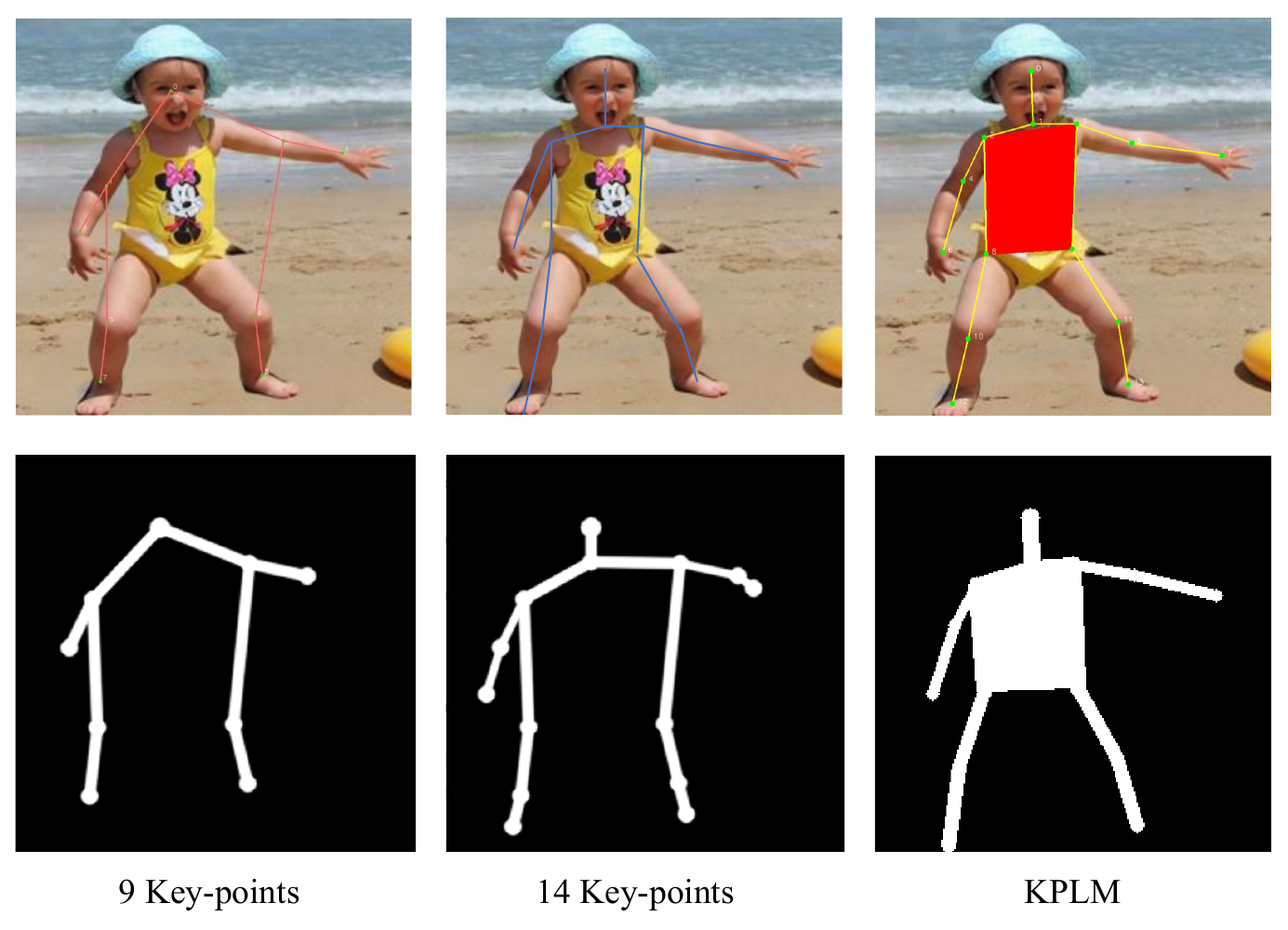}
    \caption{Example for Table.\ref{kplm}.}
    \label{ex-kplm}
\end{figure}

As shown in AL~\ref{alg-kplm}, the annotation module is designed to extract keypoint-based structural representations and generate corresponding binary skeleton masks from input images. Given an image $I \in \mathrm{R}^{256 \times 256 \times 3}$, the system records a sequence of spatial coordinates $\mathcal{P} = {\mathbf{p}i}{i=1}^k$, where each $\mathbf{p}i = (x_i, y_i) \in \mathrm{R}^2$ denotes a semantic keypoint. Upon satisfaction of a minimum cardinality constraint $k \geq k{\min}$, the coordinate set $\mathcal{P}$ is serialized into a structured annotation format (e.g., JSON) containing additional metadata such as image identifiers and pose type descriptors. Subsequently, a skeleton mask $M$ is derived by applying a mapping function $f_{\text{skeleton}}(\mathcal{P})$, which connects selected keypoint pairs according to a predefined topology. The function renders limb segments as polylines over a binary canvas using standard rasterization operations. To ensure spatial consistency, both the keypoints and the resulting mask are rescaled to align with the original image resolution. This module operates iteratively over a dataset, yielding structured supervision data for subsequent learning tasks.

\begin{algorithm}[!ht]
\caption{KPLM Annotation and Skeleton Mask Generation}
\label{alg-kplm}
\begin{algorithmic}[1]
\REQUIRE Input image $I \in \mathrm{R}^{256 \times 256 \times 3}$  
\ENSURE Keypoint set $\mathcal{P}$ and binary mask $M$

\STATE Initialize $\mathcal{P} \gets \emptyset$  
\STATE Activate annotation interface with image $I$

\WHILE{annotation session active}  
    \STATE Receive input coordinate $\mathbf{p} = (x, y)$  
    \STATE Update keypoint set: $\mathcal{P} \gets \mathcal{P} \cup \{\mathbf{p}\}$  

    \IF{$|\mathcal{P}| \geq k_{\min}$ \AND commit signal received}  
        \STATE Serialize $\mathcal{P}$ to structured format with metadata  
        \STATE $M \gets f_{\text{skeleton}}(\mathcal{P})$  
        \COMMENT{Polyline-based rendering of skeleton structure}  

        \STATE Rescale $\mathcal{P}$ and $M$ to original resolution  
        \STATE Persist $(\mathcal{P}, M)$ and load next image  
    \ENDIF  
\ENDWHILE
\end{algorithmic}
\end{algorithm}

% \begin{table}[!ht]
% \centering
% \begin{tabular}{c|c|c|c}
% %{p{1.5cm}|p{1.3cm}|p{1.8cm}|p{1.6cm}}
% \hline
%     Model & IoU $\uparrow$ & Time (ms) $\downarrow$ & Numbers $\downarrow$ \\
% \hline
%     OpenPose & 0.411 & 5547 & 17 \\
% \hline
%     MediaPipe & \textbf{0.551} & 8207 & 33 \\
% \hline
%     \textbf{KPLM} & 0.535 & \textbf{5403} & \textbf{13}  \\ 
% \hline
% \end{tabular}
% \caption{Evaluation with other human models on mask.}
% \label{kplm_mask}
% \end{table}
\section{C: More details of different limbs on STA}
In the main text, we present a comparison of the $K$ values of various parts in two images. To further demonstrate the effectiveness of our method, we will perform a unified $K$ value test and comparison on the DESOBA dataset and the DESOBAv2 dataset. The test results are shown in the Table~\ref{sta}, Table~\ref{sta1}.

\begin{figure}[!ht]
    \centering
    \includegraphics[width=7cm]{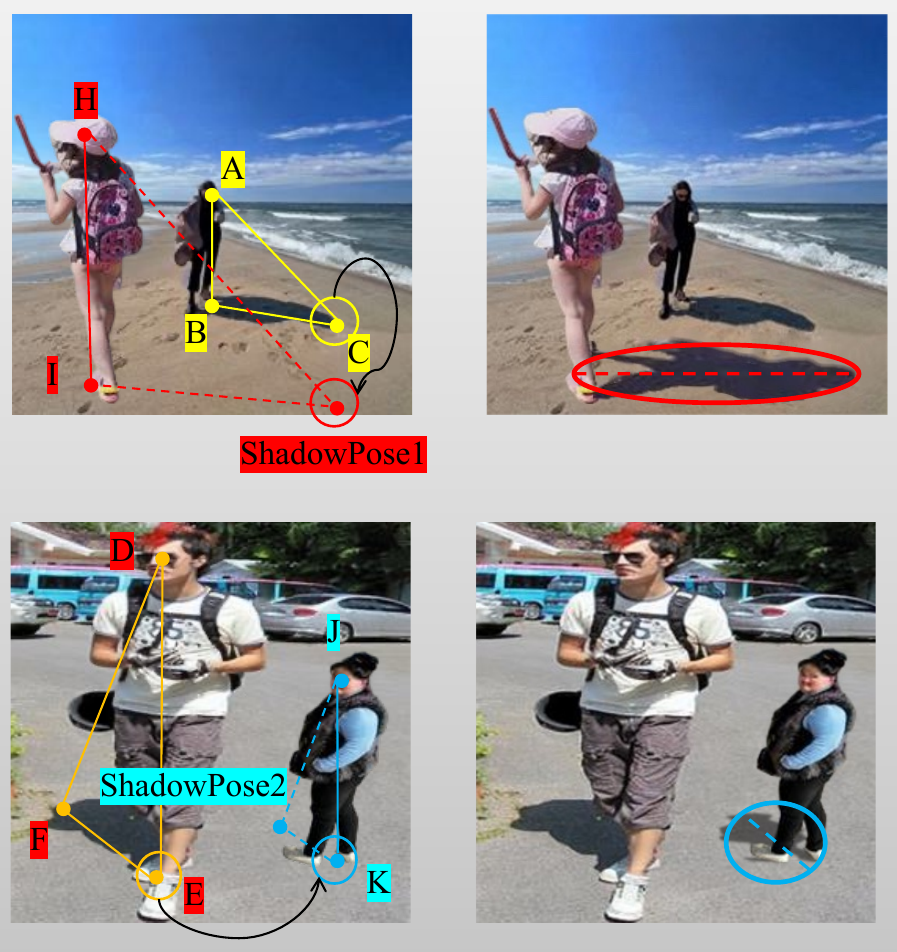}
    \caption{Example for Composite Image.}
    \label{sta-comp}
\end{figure}

We take Human (Head)* as the standard value to measure the error between the proportionality coefficient $K$ corresponding to different limbs and the standard $K$. L represents Left and R represents Right. All values are rounded to two decimal places. In the DESOBA~\cite{desoba} dataset, the average error of the value $K$ is $0.03$. In the DESOBAv2~\cite{desobav2} dataset, the error of the value $K$ is only $0.02$. Therefore, our STA is correct and effective. Additionally, in the main text, we mentioned that our method can still generate coordinates at both ends of the limb that are approximately the true value based on the proportion under the condition of no shadow prior or partial absence of limb parts. To prove the validity of this view, we also conducted experiments on real synthetic images. The visualization results are shown in the Figure.\ref{sta-comp}.

As shown in AL~\ref{alg-sta}, given a binary segmentation of the target object, we extract a representative triangle structure defined by three key vertices ${A, B, C} \subset \mathrm{R}^2$. The procedure begins by reading the input image and converting it to grayscale for further processing. A thresholding operation is applied to obtain a binary foreground mask, from which contours $\mathcal{C}$ are extracted. If no valid contours are found, the process terminates.

The dominant contour $\Gamma$ is selected based on maximal area, and a tight axis-aligned bounding box is computed to obtain its geometric parameters $(x, y, w, h)$. Vertices $A$ and $B$ are defined as the top-center and bottom-center points of the bounding box as
\begin{equation}
\label{1}
A = \left[x + \frac{w}{2},\ y \right], \quad B = \left[x + \frac{w}{2},\ y + h \right].
\end{equation}

To identify the third vertex $C$, we compute the convex hull $\mathcal{H}$ of $\Gamma$ and select a subset of candidate points located near the bottom of the object,

\begin{equation}
\label{2}
    \mathcal{P}_{\text{bottom}} = \left\{\, \mathbf{p} \in \mathcal{H} \mid p_y > y + 0.8h \,\right\}.
\end{equation}

If no such candidates exist, $C$ is assigned to coincide with $B$, resulting in a degenerate triangle. Otherwise, $C$ is chosen as the point in $$\mathcal{P}_{\text{bottom}}$$ that maximizes the Euclidean distance to $B$, resulting in extreme structural extensions

\begin{equation}
\label{3}
    C = \arg\max_{\mathbf{p} \in \mathcal{P}_{\text{bottom}}} \|\mathbf{p} - B\|_2.
\end{equation}

This geometric construction yields a triangle aligned with the vertical symmetry axis of the object, which serves as a consistent structural prior in subsequent processing stages.

\begin{algorithm}[!ht]
\caption{STA: Triangle Vertex Extraction Algorithm}
\label{alg-sta}
\begin{algorithmic}[1]
\REQUIRE Image path $p_{\text{img}}$  
\ENSURE Triangle vertices $\{A, B, C\} \subset \mathrm{R}^2$

\STATE Load image: $I \gets \texttt{ReadImage}(p_{\text{img}})$  
\STATE Convert to grayscale: $I_{\text{gray}} \gets \texttt{ToGrayscale}(I)$  

\STATE Binary segmentation and contour extraction:  
\STATE \hspace{1em} $I_{\text{bin}} \gets \texttt{Threshold}(I_{\text{gray}})$  
\STATE \hspace{1em} $\mathcal{C} \gets \texttt{FindContours}(I_{\text{bin}})$  

\IF{$\mathcal{C} = \emptyset$}  
    \RETURN $\emptyset$  
\ENDIF

\STATE Select dominant contour: $\Gamma \gets \arg\max_{C \in \mathcal{C}} \texttt{Area}(C)$  
\STATE $(x, y, w, h) \gets \texttt{BoundingRect}(\Gamma)$ \COMMENT{Bounding box} 

\STATE Define vertex $A \gets [x + \frac{w}{2},\ y]$ \COMMENT{Top center}  
\STATE Define vertex $B \gets [x + \frac{w}{2},\ y + h]$ \COMMENT{Bottom center}  

\STATE Compute convex hull: $\mathcal{H} \gets \texttt{ConvexHull}(\Gamma)$  
\STATE Extract candidate points:  
\STATE \hspace{1em} $\mathcal{P}_{\text{bottom}} \gets \{\, \mathbf{p} \in \mathcal{H} \mid p_y > y + 0.8h \,\}$  

\IF{$\mathcal{P}_{\text{bottom}} = \emptyset$}  
    \STATE $C \gets B$  
\ELSE  
    \STATE $C \gets \arg\max_{\mathbf{p} \in \mathcal{P}_{\text{bottom}}} \|\mathbf{p} - B\|_2$  
\ENDIF

\RETURN $\{A, B, C\}$
\end{algorithmic}
\end{algorithm}

\begin{table*}[htbp]
\centering
\label{bos}
\begin{tabular}{lccccccccc}
\toprule
\multirow{1}{*}{Datasets} & \multicolumn{9}{c}{DESOBA}
\\ % 合并6列
\midrule % 部分横线，仅覆盖第2-7列
Limb & Human (Head)*    & L-Elbow    & R-Elbow    & L-Wrist & R-Wrist    & L-Knee    & R-Knee    & L-Knee & R-Ankle \\ % 指标名称
\midrule
Error   & 0  & 0.03 & 0.02  & 0.04 & 0.05  & 0.05 & 0.01  & 0.03 & 0.02   \\
\bottomrule
\end{tabular}
\caption{Evaluation on DESOBA.}
\label{sta}
\end{table*}

\begin{table*}[htbp]
\centering
\label{bos}
\begin{tabular}{lccccccccc}
\toprule
\multirow{1}{*}{Datasets} & \multicolumn{9}{c}{DESOBA}
\\ % 合并6列
\midrule % 部分横线，仅覆盖第2-7列
Limb & Human (Head)*    & L-Elbow    & R-Elbow    & L-Wrist & R-Wrist    & L-Knee    & R-Knee    & L-Knee & R-Ankle \\ % 指标名称
\midrule
Error   & 0  & 0.04 & 0.01  & 0.02 & 0.03  & 0.02 & 0.03  & 0.01 & 0.01   \\
\bottomrule
\end{tabular}
\caption{Evaluation on DESOBAv2.}
\label{sta1}
\end{table*}

\section{D: Annotion}

\begin{figure}[H]
    \centering
    \includegraphics[width=8cm]{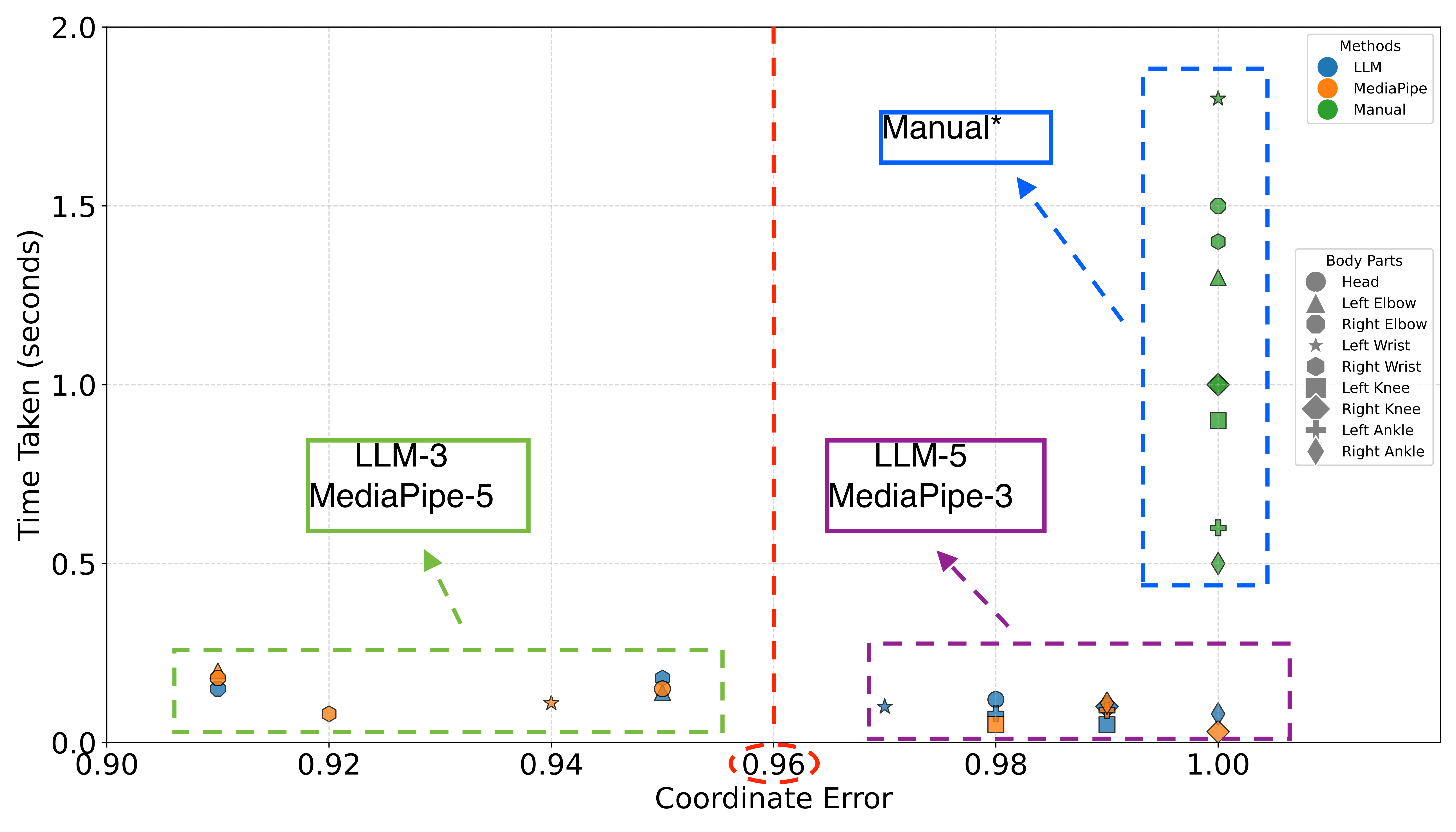}
    \caption{Annotion on different methods.}
    \label{annotion}
\end{figure}

For the dataset annotation of KPLM, we compared the accuracy for front of manual, LLM, and MediaPipe in Table~\ref{pose}. The time comparison refers to the time required for annotating a single image. Since manual annotation requires manual selection of poses and manual annotation of the position of each point, it is relatively time-consuming. However, for the annotation method of mapping MediaPipe to KPLM, due to the different number of key points, some positions may lead to inaccurate annotation, which requires manual adjustment. However, on small-order datasets, the method of LLM annotation can be adopted to perform simple shadow generation tasks. As shown in the Figure.\ref{annotion}, with $0.96$ as the midpoint and Manual as the standard, the three methods were compared. It can be seen that the LLM annotation method is superior to the MediaPipe~\cite{mediapipe} mapping method. Under similar time conditions, the accuracy of LLM is higher.
\begin{table}[!ht]
\centering
\label{pose}
\scalebox{0.7}{
\begin{tabular}{lccccccc}
\toprule
\multirow{1}{*}{Method} & \multicolumn{7}{c}{ACC} \\ % 合并6列
\cmidrule{2-8} % 部分横线，仅覆盖第2-7列
& Pose & Head     & Elbow     & Wrist     & Knee & Ankle & Time (s) \\ % 指标名称
\midrule
Manual   & 100\%  & 100\% & 100\%  & 100\%  & 100\% & 100\% & 10\\
LLM(GPT4o)      & 100\%  & 98\% & 93\%  & 96\%  & 99\% & 99\% & 1\\
MediaPipe        & 100\%  & 95\% & 91\%  & 93\%  & 99\% & 99\% & 1\\
\bottomrule
\end{tabular}
}
\caption{\textbf{Comparison of indicators using different annotation methods.} We adopts the LLM method. If only a small number of datasets are tested, manual annotation is more recommended.}
\label{pose}
\end{table}

\section{E: Post-processing GAN}
\begin{figure}[!ht]
    \centering
    \includegraphics[width=8cm]{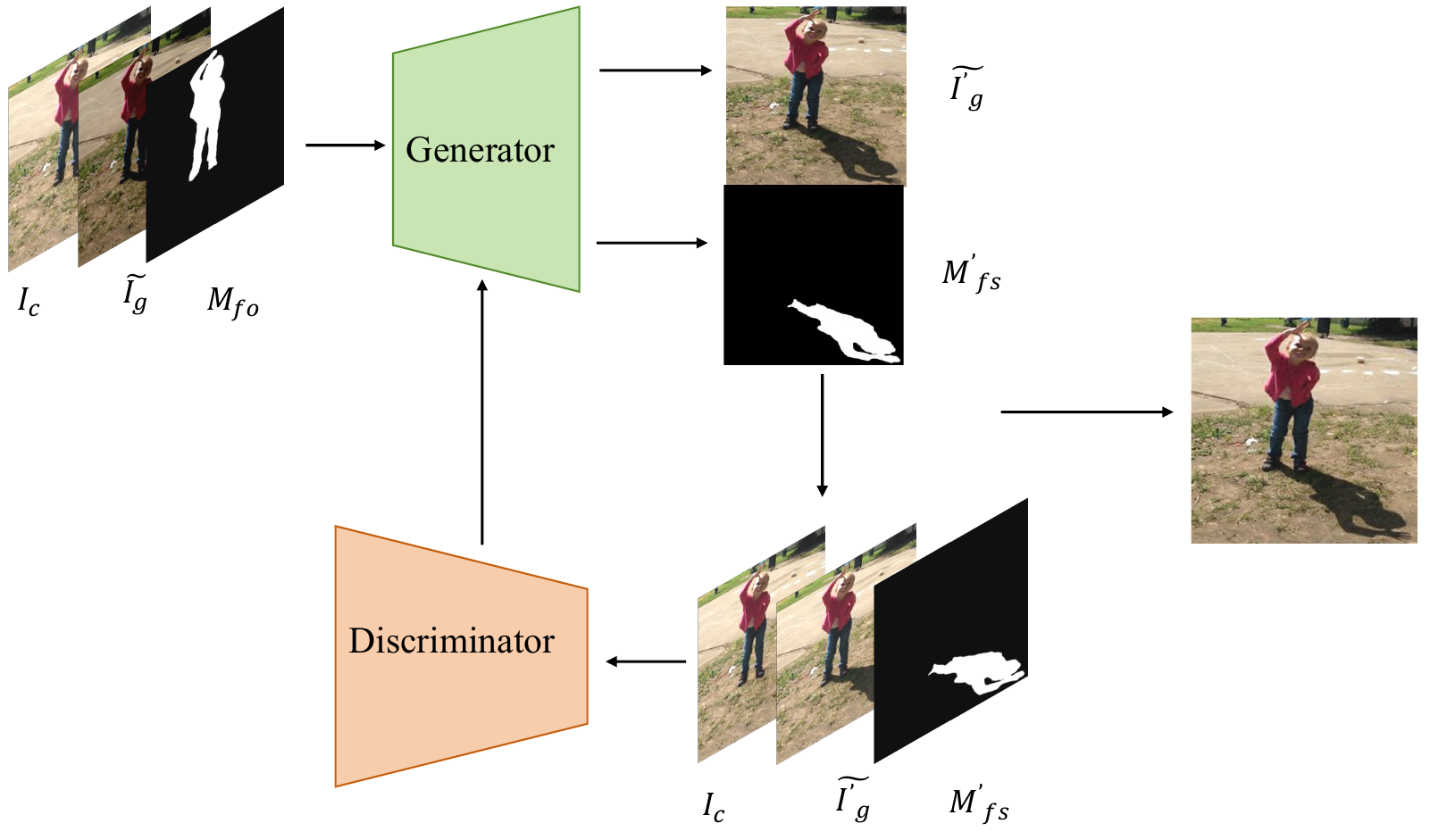}
    \caption{The framework of our post-processing network. We show our post-processing network structure which can refine a generated image.}
    \label{gan}
\end{figure}
\subsection{E1: Post-processing Network}
As shown in Figure.\ref{gan}, we adopt a GAN-based~\cite{gan} post-processing network to refine the shadow generation result from the diffusion model. Our post-processing module consists of a generator–discriminator pair trained in an adversarial manner.

The generator $G$ follows a U-Net architecture with 4 downsampling and 4 upsampling blocks. Each downsampling block consists of a convolutional layer with stride 2, followed by instance normalization and LeakyReLU. The upsampling blocks use bilinear interpolation followed by convolution, instance normalization, and ReLU. Skip connections are applied between symmetric encoder–decoder layers.

The input to the generator is the concatenation of the composite image $I_c$, the diffusion model output $\tilde{I}g$, and the foreground object mask $M_{fo}$, forming a 7-channel tensor. The generator outputs a refined shadow image $\hat{I}_g$. Our discriminator $D$ follows a PatchGAN~\cite{patchgan} architecture and is designed to distinguish real from synthesized shadow images in a local patch-wise manner. The input to $D$ is a concatenation of the composite image $I_c$, the generated (or ground-truth) shadow image ($\hat{I}_g$ or $I_g$), and the corresponding foreground shadow mask $M_{fs}$, resulting in a 7-channel tensor.

The network consists of five convolutional blocks. Each block contains a $4 \times 4$ convolutional layer with stride 2, followed by instance normalization and LeakyReLU activation. As the layers progress, the spatial resolution is reduced while the channel dimensionality increases. The final output is a patch-level score map $30 \times 30$, where each patch indicates the likelihood that the corresponding region is realistic.

We adopt the LSGAN~\cite{lsgan} objective for stable training. The adversarial loss encourages $G$ to produce realistic shadows that align with both visual quality and structural guidance from $M_{fs}$.

\subsection{E2: Loss Function}

\begin{figure*}[!ht]
    \centering
    % [width=16cm]
    \includegraphics[width=1\linewidth]{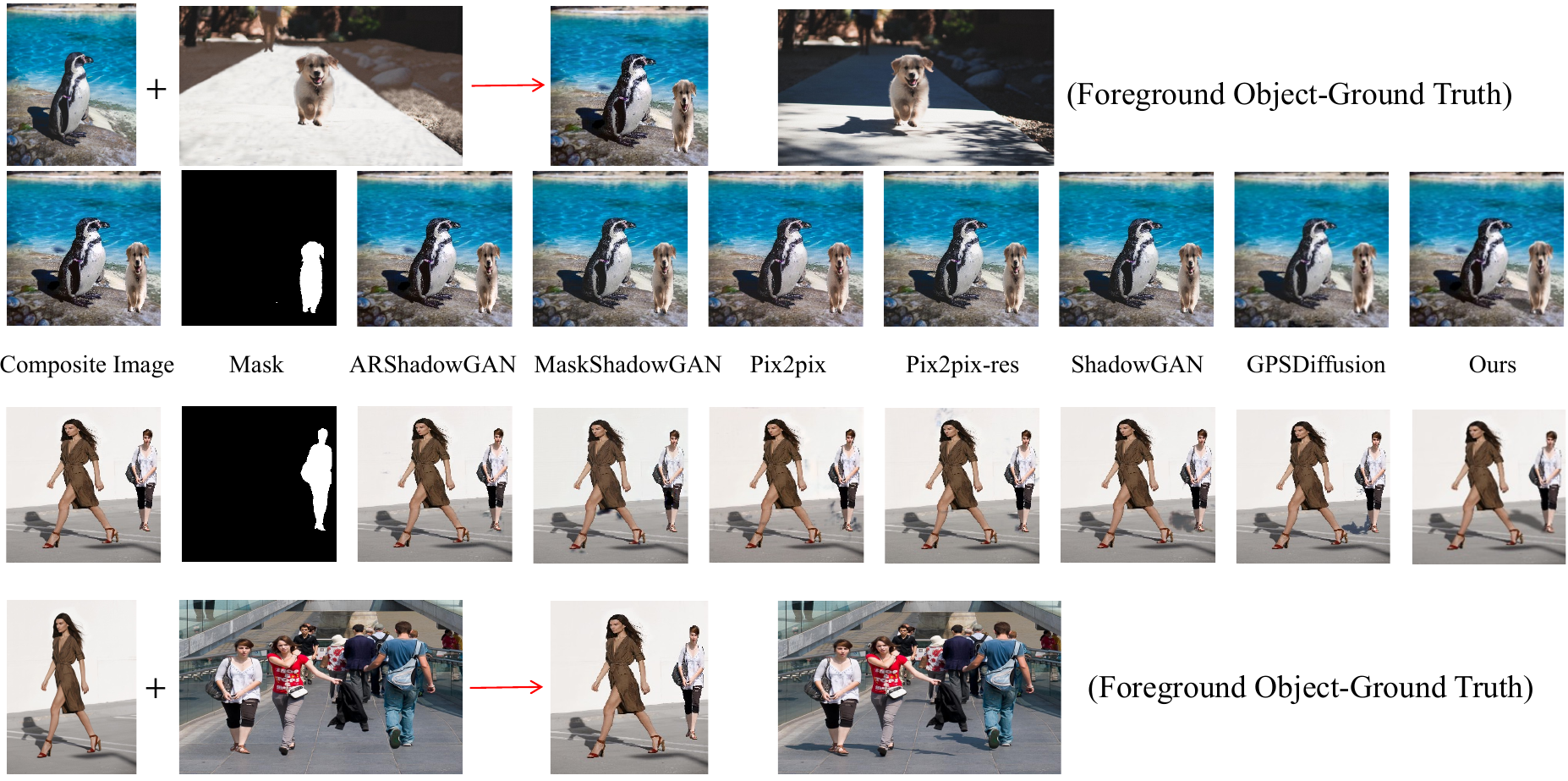}
    
    \caption{Visualization on DESOBA.}
    \label{fig:enter-label}
\end{figure*}

\begin{figure*}[!ht]
\centering  %图片全局居中
% [width=6cm,height = 6cm]
\includegraphics[width=6cm,height = 6cm]{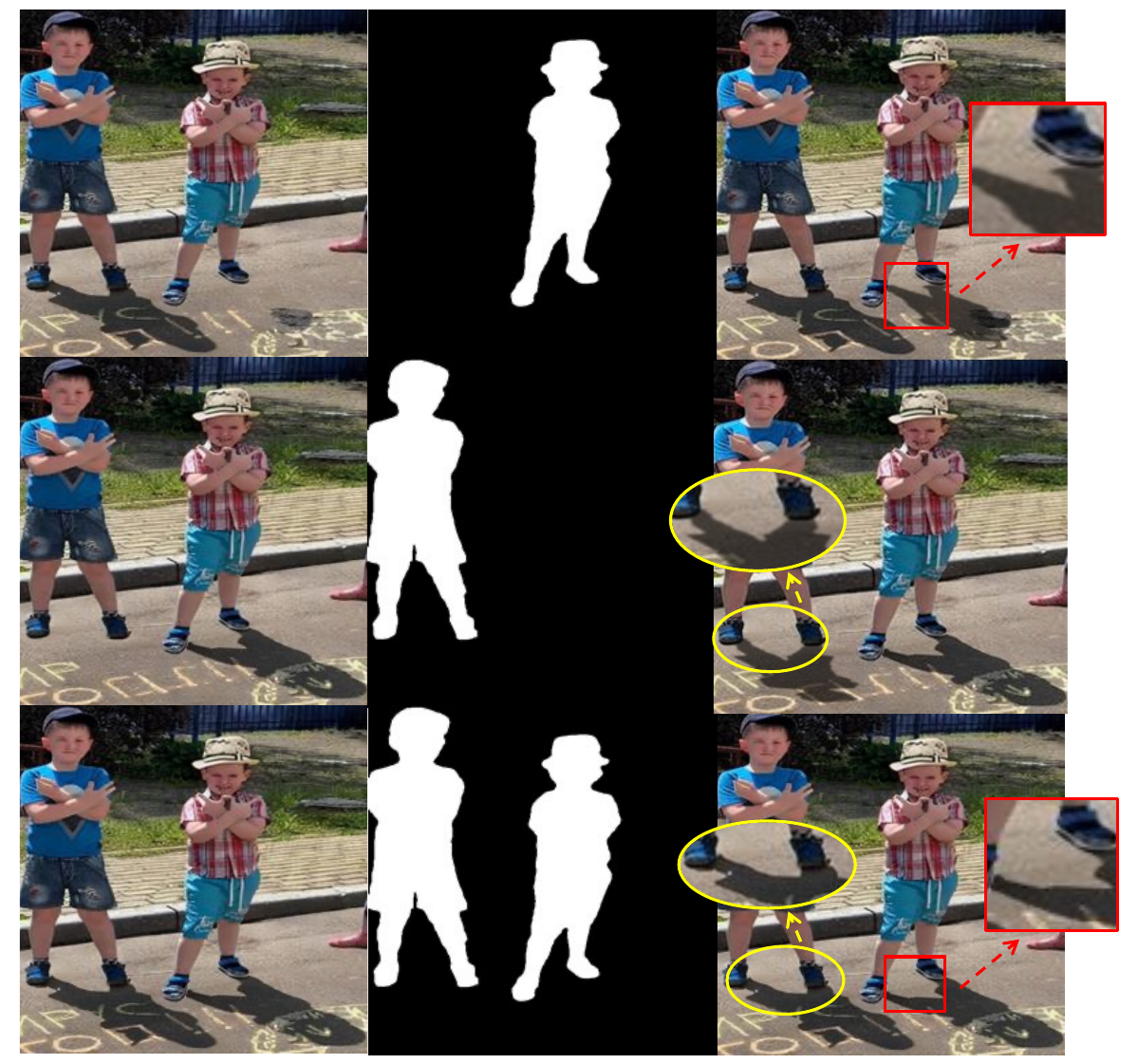}
\includegraphics[width=6cm,height = 6cm]{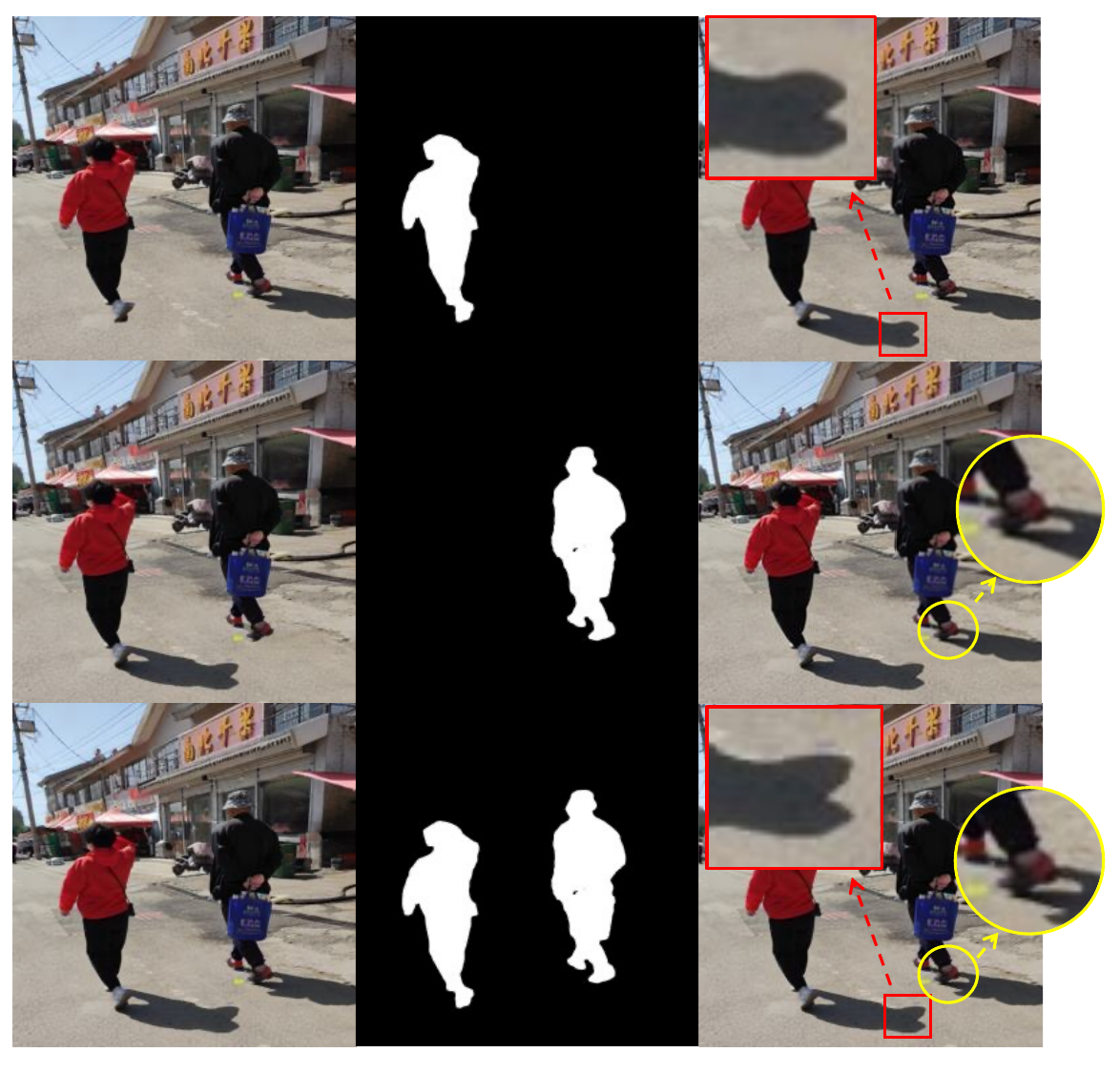}
\caption{KPLM compared with the mask of object in the most universal method OpenPose.}
\label{mask}
\end{figure*}

As shown in Figure.\ref{gan}, the generator $G$ takes the intermediate output from the diffusion model $\tilde{I}_g$ and generates a refined result $\hat{I}_g$. The discriminator $D$ is conditioned not only on the composite image $I_c$ and the generated or ground-truth shadow image ($\hat{I}_g$ or $I_g$), but also on the corresponding foreground mask $M_{fs}$. These three components are concatenated into a 7-channel tensor and fed into $D$.

We adopt the Least Squares GAN (LSGAN) loss, formulated as:

\begin{equation}
\mathcal{L}_{\text{adv}} = \left( D(I_c, \hat{I}_g, M_{fs}) - 1 \right)^2,
\label{EQ1}
\end{equation}
which encourages $G$ to generate shadows that are indistinguishable from real ones when conditioned on accurate spatial priors.

In addition to the adversarial loss, we apply an $L_1$ image reconstruction loss:

\begin{equation}
\mathcal{L}_{\text{img}} = \left\| \hat{I}_g - I_g \right\|_1,
\label{EQ2}
\end{equation}
and a perceptual loss using VGG features:

\begin{equation}
\mathcal{L}_{\text{perc}} = \sum_l \left\| \phi_l(\hat{I}_g) - \phi_l(I_g) \right\|_1.\
\label{EQ3}
\end{equation}

The complete post-processing loss is:

\begin{equation}
\mathcal{L}_{\text{gan}} = \lambda_{\text{adv}} \mathcal{L}_{\text{adv}} + \lambda_{\text{img}} \mathcal{L}_{\text{img}} + \lambda_{\text{perc}} \mathcal{L}_{\text{perc}},
\label{EQ4}
\end{equation}
where $\mathcal{\lambda}_{adv}$, $\mathcal{\lambda}_{img}$, and $\mathcal{\lambda}_{perc}$ are trade-off parameters.

\section{F: Visualization}
\subsection{F1: Visualization on DESOBA}

In the main text, we evaluated our method and existing other works through GR, LR, GS, and LS. For the shadow generation task, the most important thing is the shadow generation for synthetic images. Therefore, we compared the synthetic image generation effects of several existing methods with better performance, as shown in the following figure. This synthetic image only has background shadow guidance. There is no real shadow to guide. It can be seen that in this case, our method can still generate relatively realistic shadows, which is not available in other methods.

\subsection{F2: Visualization on Composite Image}

In the main text, we evaluated our method and the existing ones through GR, LR, GS, LS, GB and LB. And the results of single-character images were visualized. To further demonstrate the effectiveness of our method, we visualized two sets of multi-person image examples, as shown in the figure. We magnified the details of our local limbs and the same local limbs corresponding to the Ground-truth. It can be seen that our method performs well in the shadow processing of limb parts, both in terms of proportion and boundary.

\subsection{F3: Visualization with Baseline-GPSDiffusion}

GPSDiffusion~\cite{gpsdiff} is the SOTA method of 2025. As shown in the following figure, the image on the far left is generated only through the KPLM method and Diffusion for shadow generation. It can be seen that there are problems with the proportion and Angle because STA is not introduced, but the limb details and posture generation are completely correct. The $I_0-I_4$ on the right are five images generated by GPSDiffusion. It can be seen that when STA is not added, our pose and limb generation is relatively accurate.

\begin{figure}[htbp]
    \centering
    \includegraphics[width=7cm]{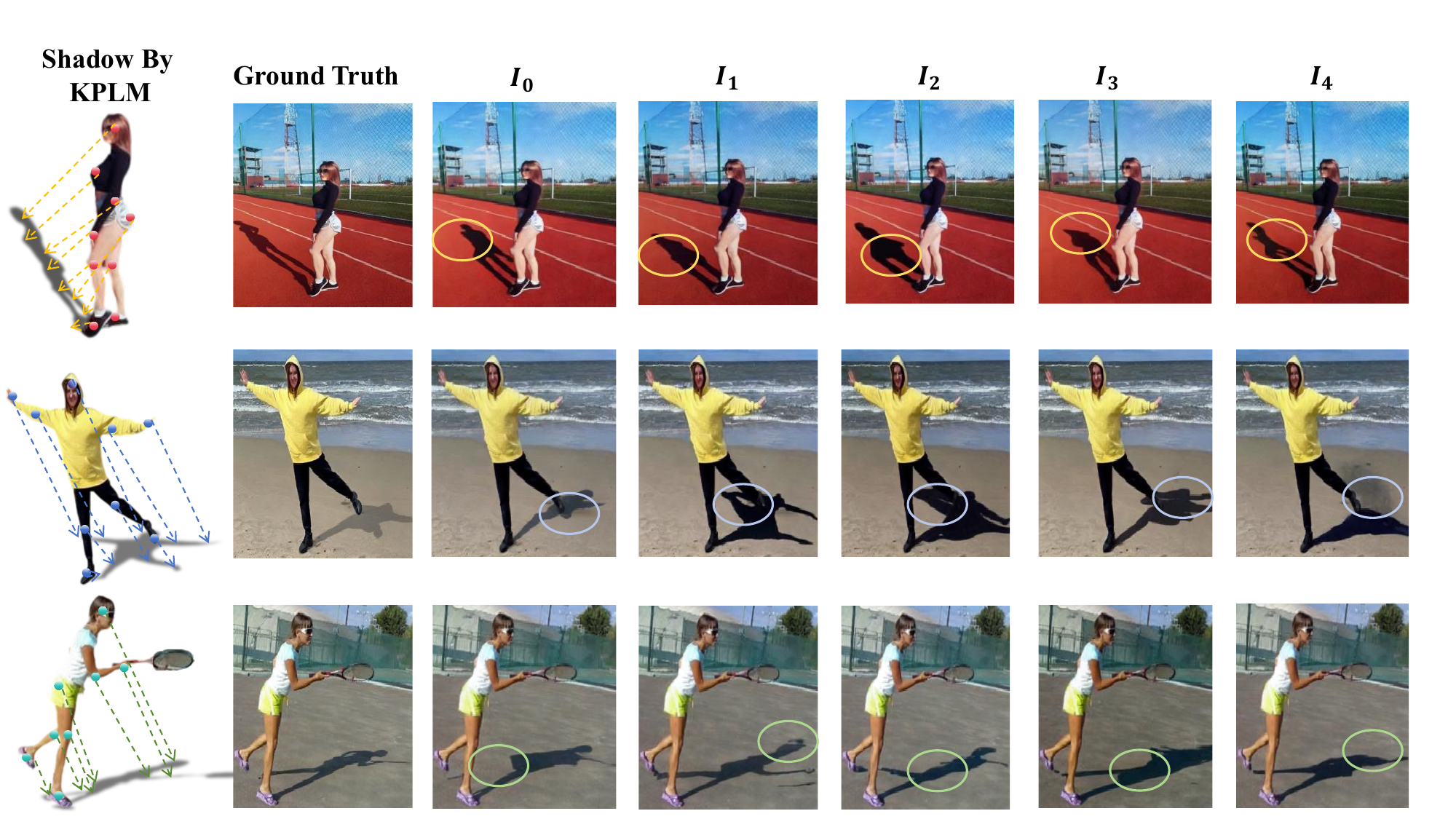}
    \caption{Visualization with Baseline.}
    \label{fig:enter-label}
\end{figure}
% \subsection{Vsiualization on different light}
% \begin{figure}[H]
%     \centering
%     \includegraphics[width=1\linewidth]{AnonymousSubmission/LaTeX/shadow with diff light.pdf}
%     \caption{Caption}
%     \label{fig:enter-label}
% \end{figure}
% Our method is designed to generate shadows for images after IC-Light Relighting, so verify that our method is useful at different light angles. Because IC-Light contains the three Light conditions of Left Light, Right Light and Botoom Light that exist in the real world when the sun shines, we visualize the effectiveness of our method in the figure.
\section{G: Limitation}
The limitation of the Diffusion model for shadow generation tasks may lie in the excessively high FLOPs required. Subsequently, more lightweight Diffusion models, such as DDSM~\cite{ddsm}, will be sought. We will also continue to explore related improvement methods and optimize the existing models.
\bibliography{aaai2026}